\documentclass{ametsocV6.2-arxiv}

\usepackage{url}
\usepackage[inline]{trackchanges}
\usepackage{soul}

\usepackage{units}
\usepackage{diagbox}
\usepackage{xparse}
\usepackage{subcaption}
\usepackage{float}
\usepackage{booktabs}

\usepackage{rotating}
\usepackage{comment}
\usepackage{amsmath}
\usepackage[overload]{empheq}
\usepackage{systeme}
\usepackage{array}
\newcolumntype{C}{>{{}}c<{{}}}
\usepackage{pifont}

\title{Drift Field Net: Learning Ocean Lagrangian advection fields from in-situ and satellite observations.}

\authors{Théo Archambault,\aff{a*}\correspondingauthor{Théo Archambault, theo.archambault@amphitrite.fr} 
Pierre Garcia,\aff{a, b*} 
Mattia Romero,\aff{c}
Anastase Charantonis,\aff{d}
Dominique Béréziat,\aff{b}
}

\affiliation{\aff{a}{Amphitrite, Paris, France}\\
\aff{b}{Sorbonne Université, LIP6, Pequan, Paris, France}\\
\aff{c}{The Ocean Cleanup, Rotterdam, Netherlands}\\
\aff{d}{INRIA, ARCHES, Paris, France}\\
\aff{*}{equal contribution}
}

\abstract{The North Pacific Subtropical Gyre (NPSG) is a major accumulation zone for floating plastic debris, resulting from basin-scale convergent ocean circulation. Effective cleanup strategies in this region rely on accurate forecasts of Lagrangian particle drift. Here, we introduce Drift Field Net (DFN), a deep neural network that predicts ocean surface flow fields from operational satellite observations. DFN is trained using a novel two-stage strategy that combines pretraining on simulated data with Lagrangian fine-tuning based on an advection-consistent loss function. This physics-informed optimization directly improves the accuracy of particle trajectory predictions.
We evaluate DFN against an operational physics-based forecasting system and demonstrate the potential of deep learning for ocean surface flow prediction. On in situ drifter trajectories, DFN reduces the mean positioning error by 20 km after a 7-day forecast compared with the operational model. Furthermore, Lagrangian fine-tuning with the proposed advection loss further reduces the positioning error by  10 km, highlighting the benefits of incorporating Lagrangian constraints into the training process.} 

\begin{document}

\maketitle

\sigstatement
Accurate forecasts of ocean surface currents are essential for predicting the transport and accumulation of floating marine debris. We introduce Drift Field Net (DFN), a deep learning framework that predicts surface velocity fields from satellite observations and is trained to improve particle trajectories through a physics-informed advection loss. By combining simulated-data pretraining with Lagrangian fine-tuning, DFN reduces seven-day drifter positioning errors by 20 km relative to an operational forecasting system, with a further 10 km improvement from Lagrangian optimization. Predicting flow fields rather than individual trajectories also enables efficient forecasting for many particles. These results demonstrate how integrating physical constraints into deep learning can improve operational ocean forecasting and support marine pollution monitoring and cleanup.

\section{Introduction}

Surface currents at the mesoscale and sub-mesoscale play a key role in regulating Earth's climate, redistributing material at the ocean surface and influencing human activity such as shipping routes. Nonetheless, ocean processes at these scales remain largely unresolved because of limited observations and computational costs. Passive tracers are physical quantities transported by the ocean currents, with variable effects from winds and waves, in a so-called advective motion. These tracers can be physical quantities, such as heat or chlorophyll concentration, but also pollutants, such as drifting plastic. The successive positions of a particle in a flow field define the Lagrangian trajectory. Predicting Lagrangian trajectories of surface particles is a key challenge for the estimation of the propagation and the concentration of pollutants in the Ocean. For instance, in the North Pacific Subtropical Gyre, floating waste from fishing and coastal activities accumulates due to basin-scale Ekman convergence ~\cite{lebreton2018}. Thus, accurate surface flow field predictions in this region are particularly important for pollutant impact assessment and upstream solutions, such as cleanup efforts.

\subsection{Observations}
A major challenge in accurately estimating ocean currents across the entire ocean lies in effectively combining information from heterogeneous data sources. This task is difficult because the observations differ substantially in terms of noise characteristics, spatial coverage, temporal sampling, and measurement precision. Amongst such data sources, satellite remote sensing provides measurements of several physical variables at a global scale. Some of the most informative variables for describing ocean surface dynamics are the Sea Surface Height (SSH) and Temperature (SST).
SSH is primarily measured using nadir-pointing satellite altimeters~\cite{DATAL3tracks}, which estimate the height of the ocean surface through active radar interferometry. Under the geostrophic approximation, SSH is closely related to ocean surface currents, as these currents result from a balance between Coriolis and pressure-gradient forces.
SST behaves as a passive tracer because surface currents transport heat. It is observed through satellite-based infrared imaging of the sea surface~\cite{DATAmur}, typically providing high spatial resolutions ranging from 1 to 4 km. However, infrared radiation is absorbed by clouds, restricting observations to cloud-free regions.
Ocean surface currents can also be measured using \textit{in-situ} drifting buoys~\cite{DATAdrifters}. These floating devices are advected by the upper layers of ocean currents, from the surface to the depth of their drogue. Some drifters are undrogued and therefore more subject to direct wind drag, similarly to some floating pollutants such as rigid macroplastics. 
By tracking their Global Positioning System positions over time, drifters provide some of the most direct and accurate observations of surface currents.

In addition to these physical quantity observations, the Automated Debris Imaging System (ADIS) \cite{de_vries_automated_2026} provides observations of afloat debris, such as plastic, observed every five seconds using GoPros$^\text{®}$ attached to vessels. This dataset doesn't provide trajectories of drifting plastic; instead, it provides sparse along-vessel trajectory measurements of the number of adrift bodies, allowing us to validate the models for the accumulation of drifting waste.

\subsection{Ocean state prediction}

Various approaches have been used to infer ocean surface circulation, primarily numerical models. Ocean Global Circulation Models (OGCM) solve a set of ocean primitive equations to represent the ocean state. Integrating the incompressible Navier-Stokes equations provides a realistic, although costly, simulation of the whole ocean system. Through data assimilation techniques, OGCMs calibrate their predicted state by leveraging observations. For example, the Global Physical Reanalysis (Glorys)~\cite{DATAglorys} uses ensemble Kalman Filters to assimilate ocean observations at 1/12° resolution on the NEMO~3.6 physical model~\cite{madec2017nemo}. To produce accurate operational forecasts, OGCMs heavily rely on their initial state, which is often hard to estimate, as it is only partially observed.  Recently, Deep Learning (DL) has emerged as one of the most promising approaches to predicting ocean states, with lower computational costs and improved performance. For instance, \cite{fablet2023} infer surface currents by assimilating SSH and SST through a physics-informed neural network trained on a physical simulation. In~\cite{martin2023, archambault_james}, the authors show how to interpolate SSH from SSH and SST observations using self-supervised learning on partial satellite observations. \cite{kugusheva2024,garcia2025} extend the approach to predicting currents, using drifting buoys as labels for a self-supervised regression forecast of current fields, thereby notably improving prediction skill. 
Other studies show that assimilating drifter observations to produce current fields is also possible. As demonstrated by research works~\cite{garcia_glofm_2026, champenois_predicting_2026}, by first learning a prior from an OGCM and then using drifter observations—and other variables for the former— through representational space optimization, neural networks can sample realistic ocean surface states.

\subsection{Particle trajectory prediction}

The state-of-the-art methods to predict particle trajectories in physical oceanography rely heavily on the accuracy of surface velocity fields, as Lagrangian trajectories are obtained through numerical advection. Open-source frameworks such as Parcels or OpenDrift~\cite{delandmeter_parcels_2019, opendrift2018} provide efficient tools for integrating particle trajectories from initial positions and velocity fields by solving the underlying advection equations.
Alternatively, DL models can be trained to predict Lagrangian trajectories directly. Several recent studies have explored this approach~\cite{della_cioppa_predicting_2025, tian_prediction_2025, botvynko2025}, employing different representations for both inputs and outputs. In all cases, zonal and meridional surface currents derived from numerical simulations constitute the primary predictors of particle motion. Additional oceanographic variables, including SSH, SST, and Sea Surface Salinity (SSS), are often incorporated to provide complementary dynamical information.
These studies formulate trajectory prediction as a Lagrangian regression problem and demonstrate that when velocity fields are imperfectly known, direct trajectory neural inference can outperform trajectories obtained through the standard advection of the estimated currents.
A central challenge in this setting arises from the mismatch between the representations of Eulerian and Lagrangian data. Ocean surface variables are naturally represented as spatial grids and are therefore well suited to convolutional neural networks (CNNs), whereas drifter trajectories consist of temporal sequences of two-dimensional positions that are less straightforward to encode and predict. To address this issue, \cite{della_cioppa_predicting_2025} and~\cite{botvynko2025} transform the initial drifter position into an image-like representation that can be processed jointly with gridded oceanographic fields. For the prediction stage, \cite{della_cioppa_predicting_2025, tian_prediction_2025} directly decode future particle positions, while~\cite{botvynko2025} estimates positions through a weighted spatial averaging procedure applied to image outputs.
Although promising, these approaches exhibit several limitations. First, trajectories are predicted individually, requiring separate network inference for each particle,  which leads to a substantially higher computational cost than conventional advection-based methods. Second, learning the mapping from Eulerian fields to Lagrangian trajectories remains challenging. As reported by~\cite{botvynko2025}, significant trajectory errors persist even when the underlying velocity fields are perfectly known, highlighting the intrinsic difficulty of directly predicting particle trajectories from gridded observations.

\subsection{Contributions}

\begin{figure}
 \centering
 \includegraphics[width=\textwidth]{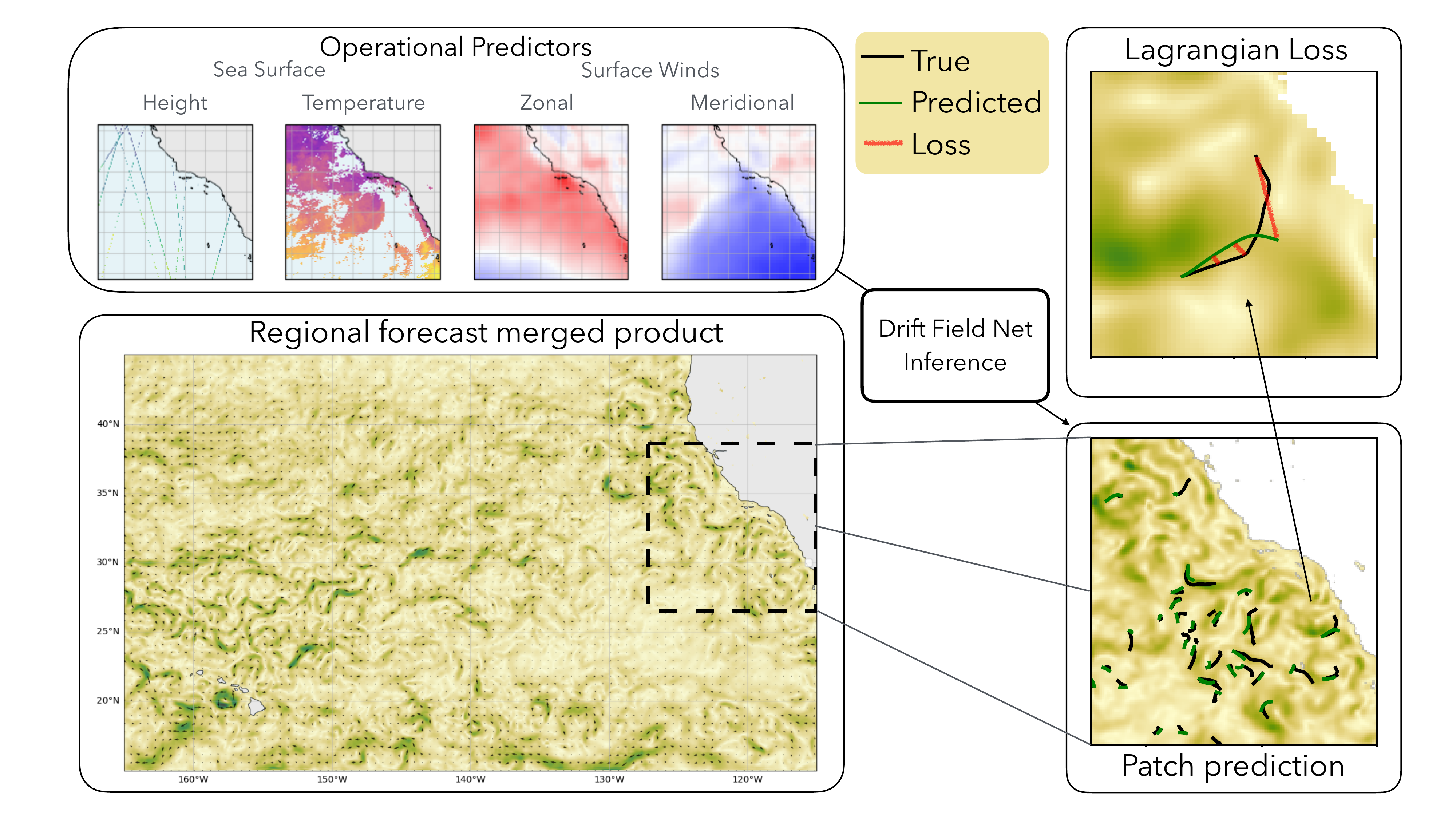}
 \caption{Illustration of the Drift Field Net method. DFN inputs patches of SSH, SST, and surface winds (top left panel) from 11 past days and outputs velocity fields for the next 7 days. These fields are then used by a differentiable numerical advection integrator, producing trajectories (bottom right panel) that are then compared to real Lagrangian drifting by the loss (top right panel). Finally, the patches are recombined to produce the current velocity on the entire area (bottom left panel).   }
 \label{fig:catchy_figure}
\end{figure}

We propose a novel deep learning framework that addresses the limitations of previous trajectory-prediction approaches. Specifically, we introduce \textit{Drift Field Net} (DFN), a neural network designed to predict a sequence of two-dimensional surface velocity fields. Unlike existing methods that directly estimate particle trajectories, DFN produces Eulerian velocity fields while being trained to optimize Lagrangian trajectory reconstruction. This is achieved through a physics-informed advection loss that compares observed and simulated drifter trajectories (Figure~\ref{fig:catchy_figure}). By operating in the Eulerian domain, DFN avoids the need to transform gridded fields into trajectory representations and generates flow fields that can subsequently be used to advect an arbitrary number of particles at negligible additional computational cost.

Second, DFN is designed as an operational forecasting system, whereas previous studies primarily focused on trajectory reconstruction or delayed-time trajectory estimation. Using past observations of SST, SSH, and surface winds, DFN predicts surface currents over a seven-day forecasting horizon.

Finally, we evaluate the proposed framework in both numerical simulations and real-world observational settings. Through a series of ablation studies and comparative experiments, we demonstrate that the proposed physics-informed loss consistently improves trajectory reconstruction across multiple evaluation metrics. We further show that the predicted flow fields can be exploited to estimate regions of plastic accumulation from \textit{in-situ} plastic observations, highlighting the practical relevance of the proposed approach for marine environmental applications.

\section{Data}

In this study, we focus on a region of the North Pacific Ocean bounded by latitudes 15°N to 45°N and longitudes 165°W to 115°W. This domain was selected because it contains the Great Pacific Garbage Patch (GPGP), a major accumulation zone where buoyant plastic debris converges under the influence of large-scale ocean currents~\cite{lebreton2018}. Over recent decades, the quantity of plastic pollution in this region has increased substantially, raising growing environmental concerns. Accurately predicting Lagrangian particle drift within this area is therefore a critical challenge, as it plays a central role in understanding plastic transport mechanisms and in supporting the planning and optimization of ocean cleanup and mitigation operations.

\subsection{Satellite data}

Following~\cite{martin2023, archambault_james, garcia2025}, we propose to leverage satellite observations to reconstruct surface ocean currents.

\textbf{Sea Surface Height (SSH).}
We employ Sea Surface Height data derived from nadir-pointing satellite altimeters and aggregated in the Level-3 multi-mission product described in~\cite{DATAL3tracks}. This dataset combines observations from multiple satellite missions that have been cross-calibrated to ensure inter-mission consistency and long-term stability. SSH is dynamically linked to surface geostrophic currents through the geostrophic balance, which describes an equilibrium between the Coriolis force and horizontal pressure gradients. Consequently, spatial gradients of SSH provide direct information on large-scale and mesoscale surface circulation, justifying its use as a primary input for current reconstruction.

\textbf{Sea Surface Temperature (SST).}
Sea Surface Temperature reflects the passive advection of heat by ocean currents. In an advective-dominated regime, thermal structures such as mesoscale eddies and frontal systems often exhibit strong temperature signatures. As shown in previous studies~\cite{resac, martin2023, archambault_james}, SST observations therefore provide indirect yet valuable information about surface circulation patterns. We utilize the Level-3 super-collated SST product described in~\cite{DATAsstl3}, which merges observations from multiple satellite platforms to produce daily global fields at a spatial resolution of 0.02°.

\textbf{Sea Surface Wind (SSW).}
Sea Surface Wind plays a fundamental role in driving wind-induced surface currents and modulating upper-ocean mixing processes, thereby influencing both temperature and salinity in the near-surface layers. We rely on the Cross-Calibrated Multi-Platform (CCMP) wind product~\cite{DATAccmp}, which integrates satellite observations (including scatterometers and passive microwave radiometers) with ERA5 atmospheric reanalysis data. This dataset provides complete global wind fields at 0.25° spatial resolution every six hours. From these data, we compute daily mean wind fields for use in our analysis.

\subsection{Drifter data}

In addition to satellite-based remote sensing technologies, \textit{in-situ} measurements are collected using buoys and other oceanographic instruments. Among these platforms, surface drifters provide direct observations of near-surface ocean currents, as they are designed to move passively with the flow. These drifters are typically equipped with a subsurface drogue extending to approximately \unit[15]{m} depth, which ensures coupling with the upper-ocean currents while minimizing the influence of direct wind forcing and wave-induced motion. As a result, surface drifters constitute one of the most reliable observational sources for characterizing surface circulation and Lagrangian ocean dynamics.
Despite their accuracy, the spatial coverage of drifters remains limited, and their distribution across the global ocean is heterogeneous. In this study, we use the data from the Global Drifter Program (GDP), provided by~\cite{DATAdrifters}. To reduce high-frequency variability, including internal waves and inertial oscillations, we preprocess the position and velocity records by applying a one-day moving average filter. We also exclude the drifters without drogues from our analysis, so that the contributions of wind and waves to the drifters' displacement can be neglected.

\subsection{Observing System Simulation Experiment}

In geosciences, a fundamental challenge lies in the fact that the true state of the system under investigation is only imperfectly known. Observational data are affected by measurement noise, incomplete spatial and temporal coverage, unobserved state variables, and sparse sampling. These limitations complicate the development and validation of robust estimation methodologies. A common strategy to address this issue is the use of an Observing System Simulation Experiment (OSSE), which emulates realistic observational errors and sampling patterns from a numerical reference simulation. Given the output of a physical model, synthetic satellite and \textit{in situ} observations can be generated, thereby providing paired datasets comprising a known “ground truth” and associated pseudo-observations. In machine learning applications to oceanography, OSSE frameworks have frequently been employed to design and evaluate data-driven reconstruction methods (e.g., \cite{archambault_james,martin2025}). Following this approach, we first assess our methodology within an OSSE framework, described below.

\textbf{Ground-truth data.}
Consistent with~\cite{archambault_james}, we use the Global Ocean Physics Reanalysis (GLORYS) dataset~\cite{DATAglorys} as the underlying reference state for our OSSE. GLORYS is based on the NEMO 3.6 ocean circulation model and provides a three-dimensional global reanalysis of key physical variables, including temperature, salinity, sea surface height (SSH), and ocean currents, at a horizontal resolution of 1/12°. The system assimilates satellite-derived SSH and SST observations, as well as \textit{in situ} drifter measurements, using a reduced-order Kalman filter. In our framework, GLORYS fields are treated as the “true” ocean state from which pseudo-observations are generated. As this reanalysis includes atmospheric forcing, we use the CCMP satellite data for all the OSSE experiments.

\textbf{SSH Level-3 pseudo-observations.}
Following~\cite{archambault_james}, we simulate along-track SSH observations by sampling the GLORYS SSH field at the locations of realistic nadir-pointing satellite altimeter tracks. The track positions are extracted from the real altimetry product described in~\cite{DATAL3tracks}, ensuring realistic spatial and temporal sampling. To emulate instrumental measurement errors, we add independent Gaussian noise with zero mean and a standard deviation of \unit[2]{cm} to the sampled values. Additional technical details regarding the OSSE configuration are provided in~\cite{DATAourOSSE}.

\textbf{SST Level-3 pseudo-observations.}
In~\cite{archambault_james}, the authors simulated optimally interpolated SST fields. In contrast, we generate Level-3 SST pseudo-observations that preserve observational gaps caused by cloud cover. Specifically, we use cloud masks derived from the dataset described in~\cite{DATAsstl3} to mask the GLORYS SST fields accordingly. Gaussian random noise with zero mean and a standard deviation of 0.5°C is then added to the remaining valid pixels to represent measurement uncertainty. This procedure yields noisy, spatially incomplete, and thus realistic SST observations.

\textbf{Drifter pseudo-observations.}
We additionally simulate Lagrangian drifter trajectories using GLORYS surface velocity fields. Drifter positions are randomly initialized on 1 January 2010, corresponding to the beginning of the study period. Their trajectories are then computed by a Lagrangian integration with a one-hour time step (see Section~\ref{sec: problem statement}). Integration continues until a drifter exits the study domain or encounters land. When a drifter leaves the domain under these conditions, it is replaced by a new drifter initialized at a random location, thereby maintaining a constant number of active drifters. This number is set to 160, which corresponds to the daily mean number of drifters present in the study region in 2024.

\section{Proposed method}

\subsection{Problem statement}\label{sec: problem statement}

Let \(F=(U,V)\) be the Eastward and Northward components of a surface Eulerian velocity field, defined as a function of any time \(t\) and position \(x,y\). Lagrangian Fluid dynamics describes how particles are passively transported by velocity fields. The trajectory of such a particle is defined by the advection equation:
\begin{equation}
    \frac{\partial x}{\partial t}= U(t,x,y)~\text{ and }~ \frac{\partial y}{\partial t}= V(t,x,y) \label{eq: adv_tep}
\end{equation}
To model this numerically, we consider that the currents are defined on a regular grid of dimension $T\times H \times W$. Therefore, evaluating the currents components at a subgrid coordinate $r=(t,x,y)$ involves interpolating the velocities at this coordinate, in our case through trilinear interpolation. The Lagrangian trajectory \(\textbf{r}\) can then be computed recursively starting from an initial time $t_0$ and position $(x_0,y_0)$ by integrating Equation~\ref{eq: adv_tep} using the Euler method:
\begin{equation}\label{eq:-advection-integration}
    r_{i+1}=
    \mathcal{A}(F,r_i)= \begin{pmatrix} t_{i}+dt 
    \\x_i+ U(r_i)
    \\y_i+ V(r_{i})
\end{pmatrix}
\text{ and } \mathbf{r}=(r_i), i\in[0,n]
\end{equation}
where $dt$ is the integration time, and $n$ the number of advection steps, and $\mathcal{A}$ the advection operator thereby defined. Alternative numerical integration schemes can also be used to solve the advection equation, such as higher-order Runge--Kutta methods. While the Euler method is only a first-order accurate integration scheme, it provides a simple and computationally efficient approximation of particle trajectories. 

\subsection{Training method}

There are several ways to train a deep learning reconstruction of velocity fields. Let $g$ be a neural network, parametrized by $\Theta$, that estimates currents from a set of inputs $X$ as following:
\begin{equation}\label{eq: nn field}
    g_\Theta(X)=\widehat{F}=(\widehat{U},\widehat{V})
\end{equation}
If we have access to ground-truth velocity fields $F=(U,V)$, for instance, when using simulation data, we can train $g_\Theta$ by minimizing a reconstruction error $\mathcal{L}$ such as the Mean Squared Error (MSE) as follows:
\begin{equation}\label{eq: eulerian loss supervised}
    \mathcal{L}( F,\widehat{F})=\frac{1}{T \times H \times W} \sum_{t,x,y}\left \|F-\widehat{F}\right\|_2^2
\end{equation}
This supervised approach has been extensively explored for ocean currents~\cite{fablet2023, kugusheva2024,wang2024}, and for SSH~\cite{archambault_james,resac}, using simulation data as a fully gridded ground truth. 

When we do not have access to complete fields but only sparse measurements, it is still possible to compute $\mathcal{L}$ at the positions where we have data.
Let $\mathbf{f}=(u_i,v_i), i\in[0,n]$ be the sparse velocity measurements of a drifter along its Lagrangian trajectory $\textbf{r}$. We can reformulate Equation~\ref{eq: eulerian loss supervised} as following:
\begin{equation}\label{eq: eulerian loss unsupervised}
    \mathcal{L}( \mathbf{f},\widehat{F})=\frac{1}{n} \sum_{i}\left \|f_i-\widehat{F(r_i)}\right\|_2^2
\end{equation}
This self-supervised approach allows training even without a fully gridded ground truth, which corresponds to most of the cases when using real measurements of velocity. Several works used this training strategy to estimate currents without simulation training \cite{garcia2025} and SSH \cite{martin2023,archambault_james}.
As the losses described in Equation~\ref{eq: eulerian loss supervised}~and~\ref{eq: eulerian loss unsupervised} compute the current distance for each position independently, in the following, we name it the \textit{Eulerian loss}. 

Another training approach is to predict the position of a drifting particle through time, instead of a velocity field. In \cite{botvynko2025}, the authors introduce DriftNet, a network able to transform a velocity field into a timeseries of Lagrangian positions as follows:

\begin{equation}\label{eq: loss driftnet}
        g_\Theta(X)=\widehat{\mathbf{r}} \text{ ~~~and~~~ }
    \mathcal{L}( \mathbf{r},\widehat{\mathbf{r}})=\frac{1}{n} \sum_{i}\left \|r_i, \widehat{r_i}\right\|_\text{hav}
\end{equation}

In this study, we present a new training method that mitigates DriftNet's limitations, while still outperforming Eulerian training on Lagrangian metrics. We propose a neural network that outputs an Eulerian field (following Equation~\ref{eq: nn field}), although it was trained using an advection-informed loss. Given an estimation of the velocity field $\widehat{F}$, we start from a position $r_0$, and recursively apply the advection integration defined in Equation~\ref{eq:-advection-integration} to generate a drift trajectory, namely, subsequent $(\widehat{r_i}), i\in [1,n]$. Then, the simulated trajectory is compared to the true one as follows:
\begin{equation}\label{eq: advection loss}
        \mathcal{L}( \mathbf{r},\widehat{F})=\frac{1}{n} \sum_{i=1}^n\left \|r_i, \mathcal{A}^{(i)}\left(\widehat{F},r_0\right)\right\|_\text{hav}
\end{equation}
where $\mathcal{A}^{(i)}\left(\widehat{F},r_0\right)=\widehat{r_i}$ is the estimated position of the particle after $i$ advection steps in the flow $\widehat{F}$ from position $r_0$, as defined in Equation~\ref{eq:-advection-integration} with $r_0$ (ground-truth) as the initial position.
In practice, we often have multiple ground-truth trajectories for the same example; in that case, we average the loss over the available trajectories and take the SGD step on this average. It's also possible that a drifter exits the example area; in that case, we compute the drift up to the moment when the simulated particle exits the area and compute the loss on that smaller trajectory, effectively decreasing $n$. If the true particle exits the area but not the predicted one, we continue to optimize the advection loss.

\subsection{ORCAst model}

We use the ORCast model introduced in \cite{garcia2025}, which is derived from the SimVP architecture~\cite{zhangyang_simvp_2022} to forecast seven days of currents. The neural network inputs satellite observations of SSH, SST, and the two wind components between timesteps $0~\rightarrow~T$ and outputs the two current components between timesteps $T+1~\rightarrow~T+\tau$ as in: 
\begin{equation}
    g_\Theta\left(X^{0\rightarrow T}\right)=\widehat{U}^{T+1\rightarrow T+\tau},\widehat{V}^{T+1\rightarrow T+\tau}
\end{equation}
where $X$ is the input of SSH, SST, and wind; in the following, we take 12 days of input observations $(T=11)$ to estimate 7 days of forecast ($\tau=7)$. 

\begin{figure}[htpb]
    \centering
    \includegraphics[width=0.8\linewidth]{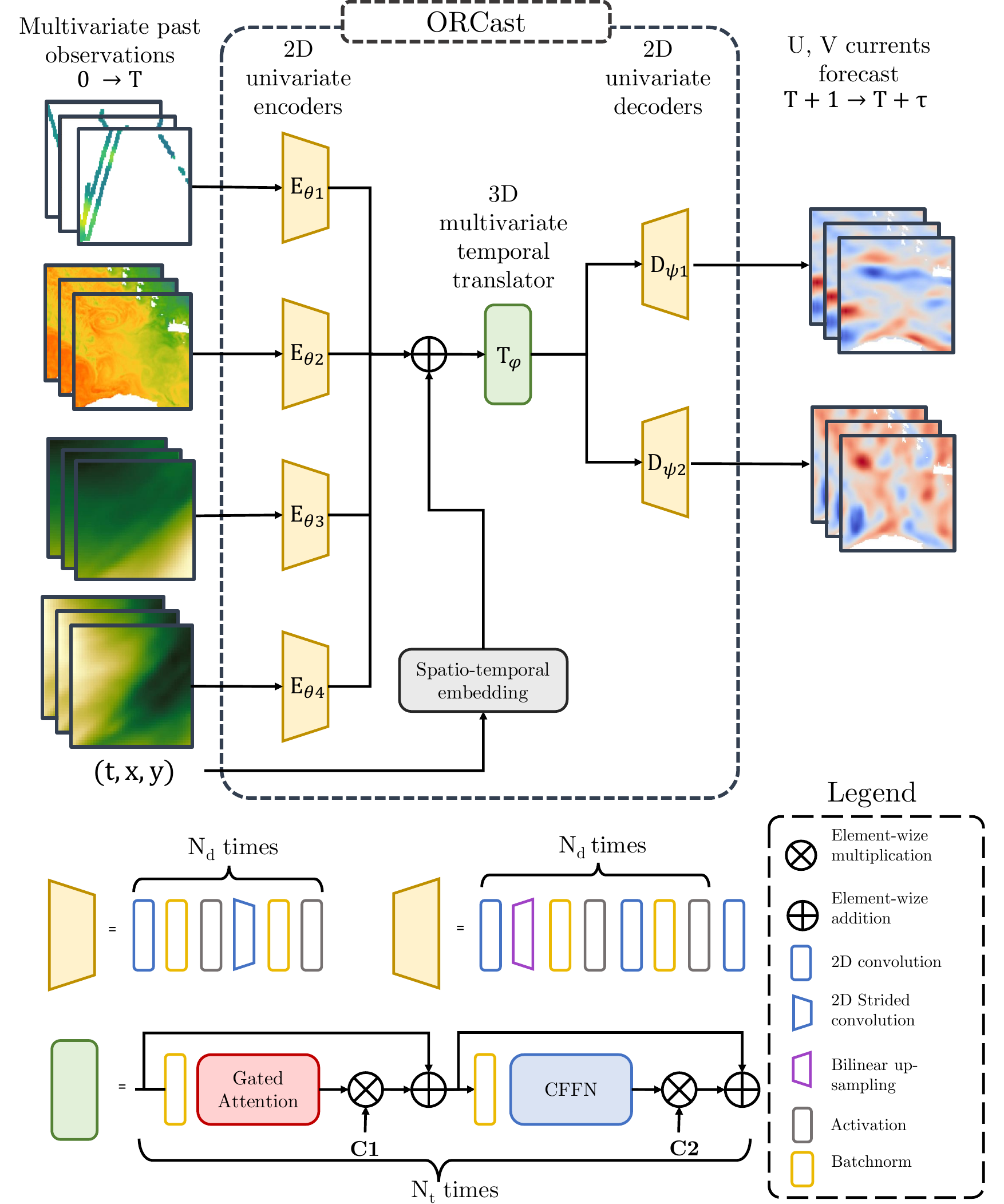}
    \caption{Representation of the model architecture. The main differences with ORCast are that we have added two wind encoders, and that we produce currents directly, without any SSH decoder.}
    \label{fig: model architecture}
\end{figure}

The different input variables are first passed to spatial encoders ($E_\theta$ in Figure~\ref{fig: model architecture}), which share the same architecture but have specific weights, one for each encoded variable $(\theta_1,\theta_2,\theta_3,\theta_4)$. Each timestep is encoded separately, but using the same weights. 
Then the obtained latent spaces are combined by a latent temporal encoder $T_\phi$ that aggregates the information of all the input timesteps from the different variables as well as a spatial-temporal positional encoding, informing the network of the patch position on the globe. 
Finally, two decoders $D_\psi$ output the two current components separately. As for the encoders, the decoders share their architecture, but with different weight sets $(\psi_1,\psi_2)$ and decode each timestep separately. The final weights of the network is the aggregation of all the weights of the different blocks $\Theta=(\theta_1,\theta_2,\theta_3,\theta_4,\phi,\psi_1,\psi_2)$.

The size of the spatial patch is $400 \times 400$ grid points at a spatial resolution of $1/30$°, corresponding to an approximate spatial extent of 13.33° per patch (i.e., roughly 1400 km). 
For additional architectural and implementation details, see~\cite{garcia2025}. 

The model takes as input 12 consecutive days of observations (from day -11 to day~0) and produces a 7-day daily forecast.
Inputs are provided as spatial patches of size $400 \times 400$ grid points at a spatial resolution of $1/30$°, corresponding to an approximate spatial extent of 13.33° per patch (i.e., roughly 1400 km). 
For additional architectural and implementation details, we refer the reader to \cite{garcia2025}.

\subsection{Training details}

\textbf{Dataset split.} In all experiments, we separate train, validation, and test periods. We train our model using data from 2010 to 2022, validate the hyperparameters using 2023 data, and reserve 2024 data for test validation, on which we will conduct the current analysis.

\textbf{Patch combination. } ORCAst is a patch-based model that estimates the currents inside patches of side $13.3^\circ$. To combine predictions for the entire region, we use the Gaussian Kernel Weighted Average (GKWA) strategy described in \cite{martin_deep_2024}, merging patches using a Gaussian Kernel for the predictions in overlapping patch regions, smoothing patch border artifacts. Thus, we apply GKWA on patches of side $13.3^\circ$ with a stride of $10^\circ$, using the patch-centered Gaussian kernel with $\sigma\simeq1^\circ$.

\section{Experiments and results}
In the following, we propose two sets of experiments. The first one is a conceptual run based on our Glorys OSSE, which uses an underlying ground truth to demonstrate the feasibility and the interest of our approach. The second one is a demonstration of the operational potential of the method, which is applied to real-world data and compared to the operational forecast system.

\subsection{OSSE Experiments}

In this section, we compare Eulerian and Lagrangian training strategies within the GLORYS OSSE framework. Three training configurations are evaluated:

\begin{itemize}
\item \textbf{Eulerian:} The model is trained to predict ground-truth velocity fields from GLORYS. The loss function corresponds to a fully supervised mean squared error (MSE) applied over the entire field (see Equation~\ref{eq: eulerian loss supervised}).

\item \textbf{Eulerian + Lagrangian:} The model is first pretrained using the Eulerian approach described above, and subsequently fine-tuned using a Lagrangian loss computed from simulated drifter trajectories (see Equation~\ref{eq: advection loss}).

\item \textbf{Lagrangian:} The model is trained exclusively using the Lagrangian loss, without any prior Eulerian pretraining.

\end{itemize}
We provide a summary of the different training configurations in Table~\ref{tab: training stages}.

\subsubsection{Lagrangian Particle Distance}

The first evaluation conducted in this OSSE framework assesses the ability of the different models to accurately predict the positions of Lagrangian drifters after seven days of advection. To this end, a seven-day forecast is initialized for a given day using the initial positions of simulated drifters. Predicted trajectories are then obtained using the advection solver defined in Equation~\ref{eq:-advection-integration}. This procedure provides both the ground-truth drifter positions after seven days (as simulated within the GLORYS velocity fields) and the corresponding model-based estimates.

Two performance metrics are considered. The \textit{Lagrangian Particle Distance} (LPD) is defined as the haversine distance between the predicted and true drifter positions after seven days. The \textit{Normalized Lagrangian Particle Distance} (NLPD) is defined as the ratio of the LPD to the total length of the true trajectory. While the LPD provides an absolute error in kilometers, independent of the trajectory length, the NLPD normalizes the error, enabling comparisons across trajectories of different lengths.

We report the performance of the different training strategies in Table~\ref{tab: drift_metrics_osse}. For reference, we also include a baseline denoted as \textit{No motion}, which corresponds to assuming no displacement, \textit{i.e.}, comparing the final drifter position to its initial position.

The results indicate that the Eulerian model outperforms the purely Lagrangian model. This outcome is expected, as the Eulerian approach is trained on fully gridded velocity fields, whereas the Lagrangian model relies solely on sparse trajectory data, which contains less spatial information. However, the Eulerian model, fine-tuned with the Lagrangian loss, achieves the best performance for both the LPD and NLPD. This result suggests that incorporating Lagrangian information during fine-tuning improves trajectory prediction accuracy.

This finding highlights that Eulerian and Lagrangian objectives do not minimize the same error structures. In particular, minimizing trajectory-based errors such as the LPD requires accounting for the accumulation and propagation of velocity errors over time through the advection process.

Although the NLPD follows similar trends to the LPD, its error remains relatively high. For instance, the mean NLPD of the Eulerian + Lagrangian model reaches approximately 60\% of the trajectory length after seven drift days. This can be attributed to the presence of short and less predictable trajectories, for which surface current signatures are weaker. Additionally, the Lagrangian training loss does not explicitly normalize errors by trajectory length, which may bias the optimization toward reducing errors on longer trajectories, where absolute distances are larger.

\begin{figure}[htbp]
    \centering
        \includegraphics[width=\linewidth]{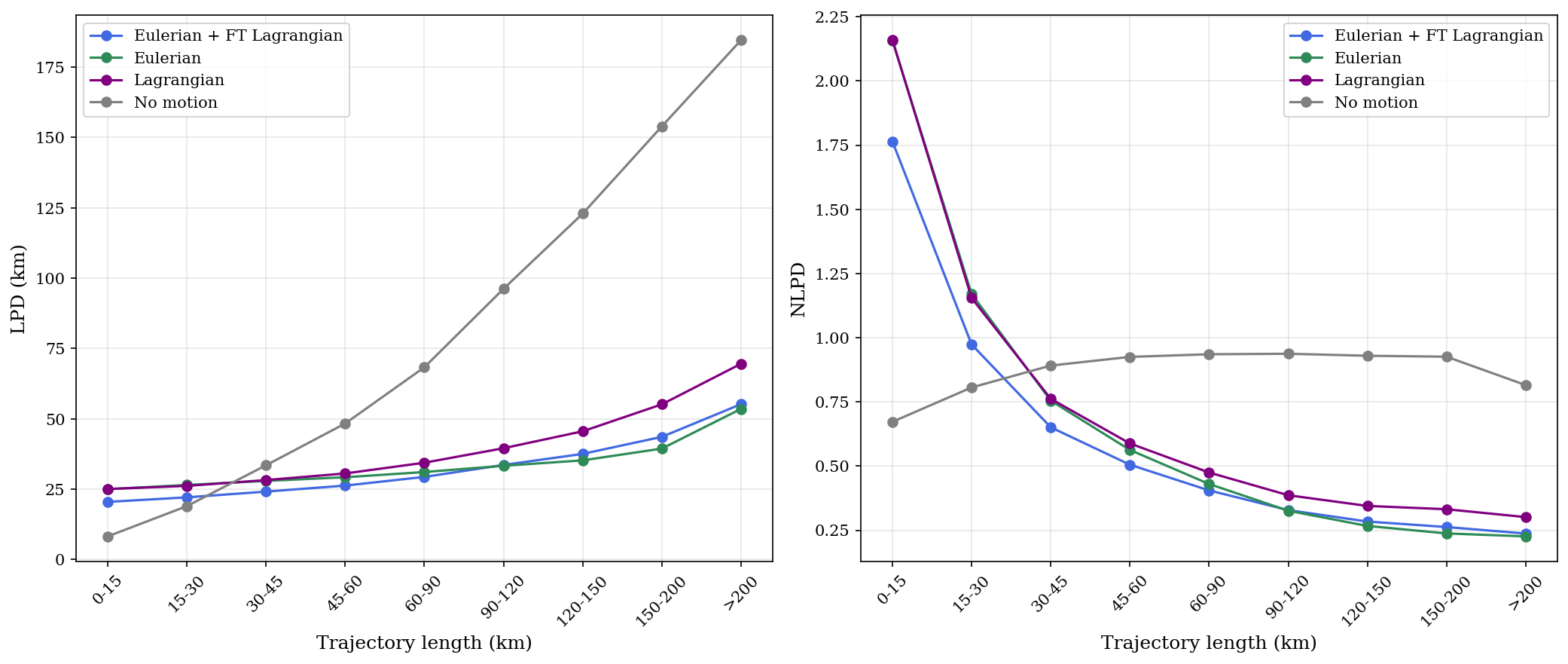}
        \caption{LPD and NLPD as a function of trajectory length on the OSSE.}
    \label{fig: error_vs_trajectory_length}
\end{figure}

Figure~\ref{fig: error_vs_trajectory_length} further illustrates the dependence of LPD and NLPD on the true trajectory length. All models exhibit similar behavior: for short trajectories, relative errors are large, whereas for longer trajectories (exceeding \unit[200]{km}), the positioning error stabilizes at approximately 25\% of the trajectory length.

\begin{table}[htbp]
\centering
\small
\begin{tabular}{l|c|c|l}
\toprule
Training stage & Inputs & Targets & Loss  \\
1 OSSE Eulerian & $\text{SST}_{\text{sim}}$, $\text{SSH}_{\text{sim}}$, Wind & $\text{UV}_{\text{sim}}$& Full field Eulerian MSE (Equation~\ref{eq: eulerian loss supervised})\\
2 OSSE Lagrangian & $\text{SST}_{\text{sim}}$, $\text{SSH}_{\text{sim}}$, Wind & $\text{Drifters}_{\text{sim}}$& Lagrangian particle loss (Equation~\ref{eq: advection loss})\\
3 Real-world Eulerian & $\text{SST}_{\text{sat}}$, $\text{SSH}_{\text{sat}}$, Wind & $\text{Drifters}_{\text{sat}}$& Sparse field Eulerian MSE (Equation~\ref{eq: eulerian loss unsupervised})\\
4 Real-world Lagrangian & $\text{SST}_{\text{sat}}$, $\text{SSH}_{\text{sat}}$, Wind & $\text{Drifters}_{\text{sat}}$& Lagrangian particle loss (Equation~\ref{eq: advection loss})\\
\hline
\end{tabular}
\caption{Different training stages}
\label{tab: training stages}
\end{table}

\begin{table}[htbp]
\centering
\begin{tabular}{lcccc}
\toprule
 & \multicolumn{2}{c}{LPD (km)} & \multicolumn{2}{c}{NLPD}   \\
\cmidrule(lr){2-3} \cmidrule(lr){4-5}  
Method & Mean & Std & Mean & Std  \\
\midrule
No motion & 52.7 & 31.5 & 0.89 & \textbf{0.14}\\
DFN Eulerian & 29.6 & 19.7 & 0.68 & 0.65\\
DFN Lagrangian & 31.9 & 20.7 & 0.71 & 0.62\\
DFN Eulerian + FT Lagrangian & \textbf{27.1} & \textbf{17.6} & \textbf{0.60 }& 0.52\\\bottomrule
\end{tabular}
\caption{Lagrangian Particle Distance Drift metrics on the OSSE, after seven days of drift.}
\label{tab: drift_metrics_osse}
\end{table}

\subsubsection{Spectral analysis}\label{sec:spectral_osse}

\begin{figure}
    \centering
    \begin{subfigure}{1\textwidth}
        \centering
        \includegraphics[width=\linewidth]{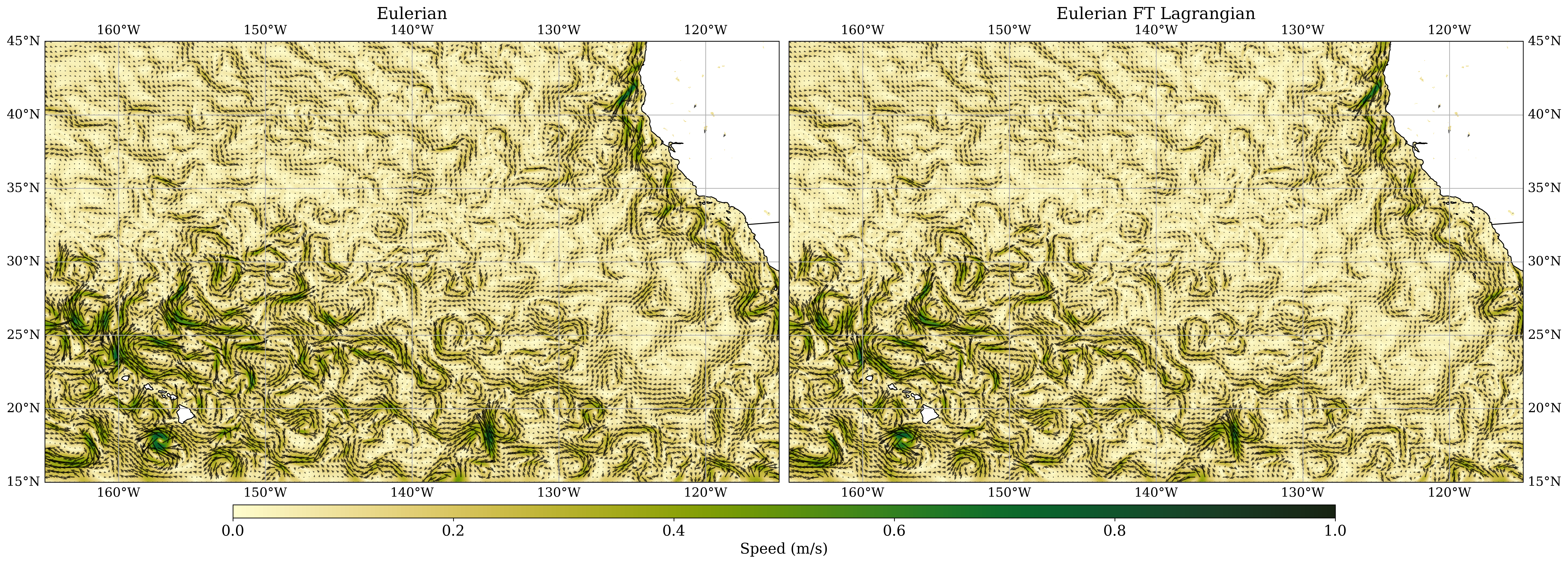}
        \includegraphics[width=\linewidth]{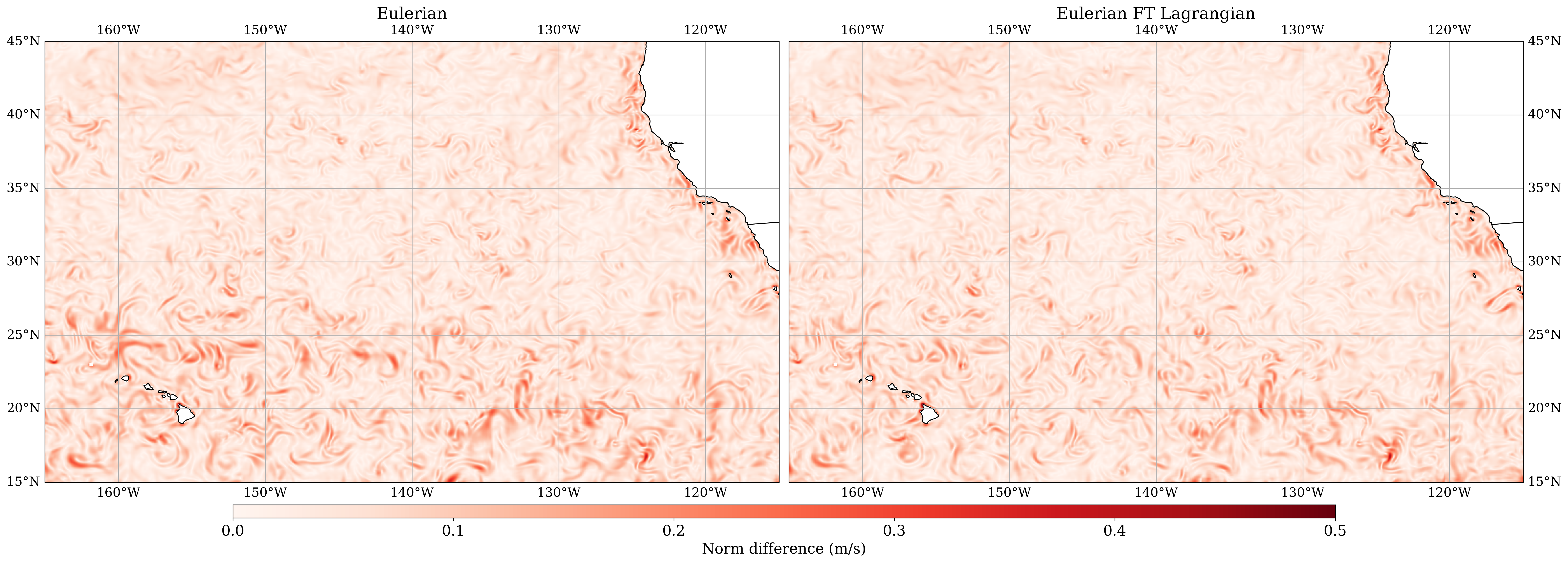}
    \end{subfigure}
    \caption{Magnitude of currents and errors on the OSSE on 6 June 2024.}
\label{fig: currents plot}
\end{figure}

We define the PSD of the currents $F$ as :
\begin{equation}
    \operatorname{PSD}_F(\omega)=\frac 1 {N_\text{lat}} \sum_i^{N_\text{lat}}\frac{\left| \widetilde{F}_i(\omega)\right|^2}{\Delta \text{lon}_i} \label{eq:PSD}
\end{equation}
, with $\omega$ the frequency, $\widetilde{F}_i$ the discrete one dimensional Fourier transform of $F$ at the $i$-th row, $N_\text{lat}$ the number of rows, $\Delta \text{lon}_i$ the kilometer longitudinal spacing associated with the $i$-th row, and $\left| \cdot \right|$ the complex modulus. Not all rows are land-free. In a row, when undefined current values are present, we compute the maximal segments from all connected regions and retain them for computation if they contain the longest wavelength. Then, for each frequency of interest, average the squared modulus of the kept segments.

\begin{figure}
    \centering
    \begin{subfigure}{.48\textwidth}
        \centering
        \includegraphics[width=\linewidth]{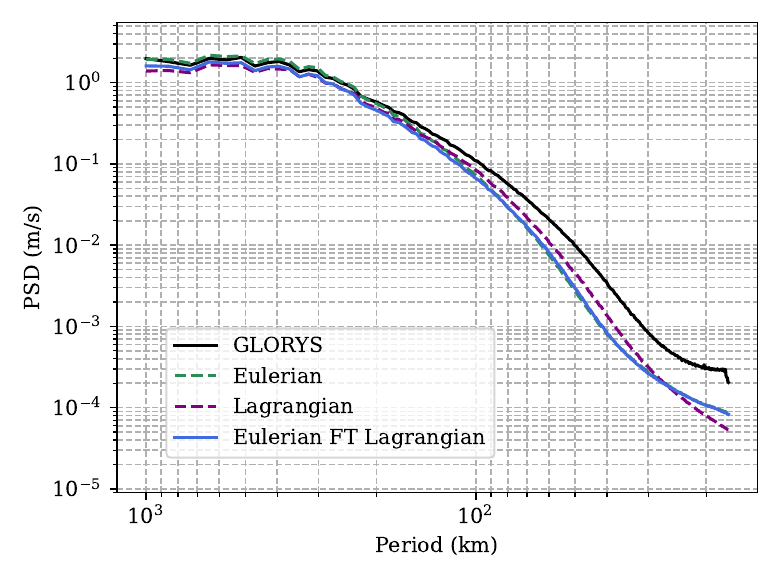}
        \caption{First day of the forecast.}
    \end{subfigure}
    \begin{subfigure}{.48\textwidth}
        \centering
        \includegraphics[width=\linewidth]{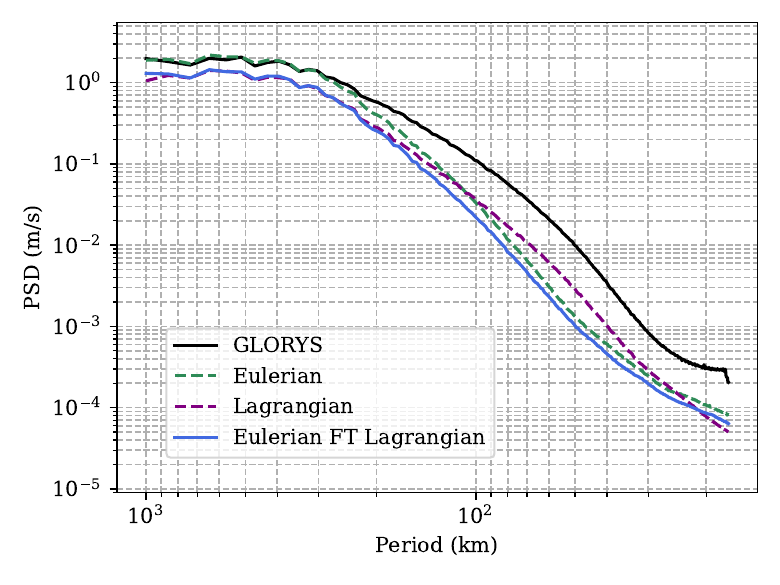}
         \caption{Seventh day of the forecast.}
     \end{subfigure}
     \caption{Year 2024 average PSD of the three OSSE models, and the GLORYS target.}
    \label{fig:OSSE_PSD}
\end{figure}

Spectral activity quantification is the cornerstone of evaluations of geophysical realism. Thus, we are interested in the evolution of the Power Spectral Density (PSD) equation~\ref{eq:PSD} with respect to the training objectives for the DL models. Since we study velocities, we use complex numbers as they are compatible with 2D vector descriptions. We compare the PSD of currents with $F=U +iV$ in Figure~\ref{fig:OSSE_PSD}.

Figure~\ref{fig:OSSE_PSD} depicts a general decrease in the spectral activity for the \textit{Lagrangian} and \textit{Eulerian FT Lagrangian} models compared to the \textit{Eulerian} model. This is especially true for the seventh day forecast and the large wavelengths for the \textit{Eulerian FT Lagrangian} model.

Furthermore, assessing the spatial scale-dependent errors of predicted fields is key to understanding how modeling inaccuracies arise. To do so, we compute the effective resolution derived from the PSD similarly to \cite{ODC2021a}. The authors define the effective resolution of a model as the smallest resolved wavelength. Formally, it is the wavelength above which all wavelengths yield error amplitudes that remain below half of the ground-truth signal amplitude.

We compare the PSD of the error $F-\hat{F}$ with $F$, the ground-truth velocity fields, $\hat{F}$, the predicted velocity, providing a metric that maps the currents' error as a function of spatial scales. We define the Power Spectral Reconstruction Ratio (PSRR) as follows:
\begin{equation}\label{eq: PSRR}
    \text{PSRR}\left( \hat{F}, F\right)=1-\frac{\text{PSD}( \hat{F} - F)}{\text{PSD}(F)}
\end{equation}

\begin{figure}
    \centering
    \begin{subfigure}{.48\textwidth}
        \centering
        \includegraphics[width=\linewidth]{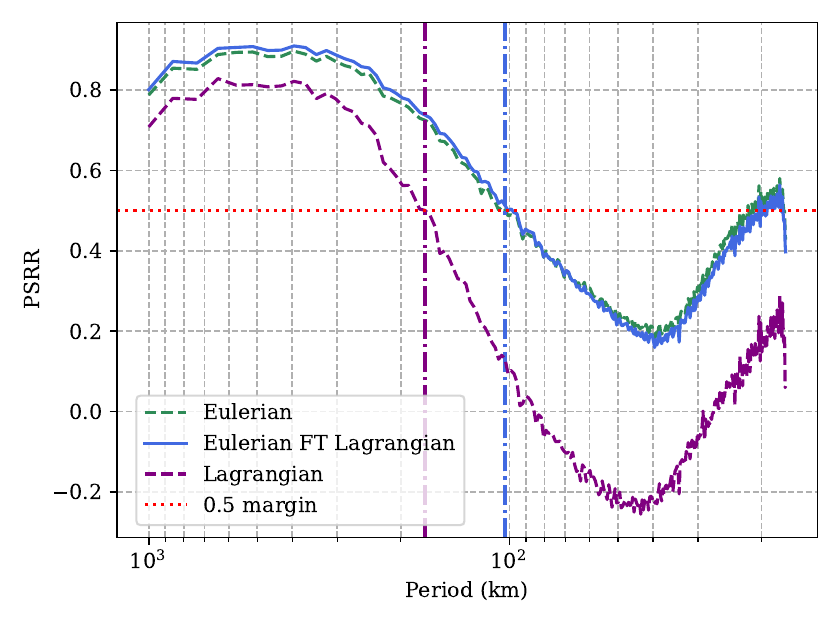}
        \caption{First day of the forecast.}
    \end{subfigure}
    \begin{subfigure}{.48\textwidth}
        \centering
        \includegraphics[width=\linewidth]{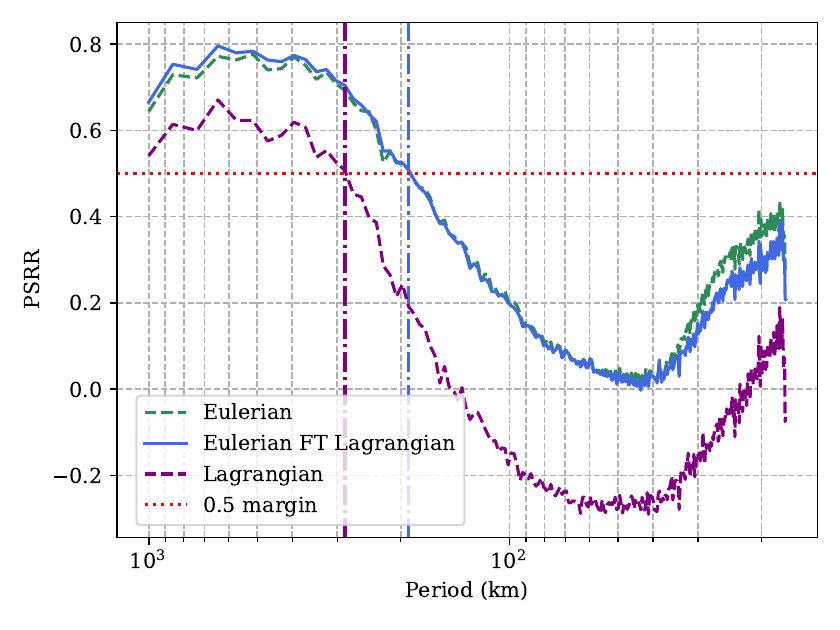}
        \caption{Seventh day of the forecast.}
    \end{subfigure}
    \caption{Year 2024 average of currents PSRR with respect to the ground-truth (GLORYS) signal.}
    \label{fig:OSSE_effective_resolution}
\end{figure}

To ensure the validity of spectral computations, the model predictions are first linearly interpolated from a 1/30° grid to a 1/12° grid, matching the resolution of GLORYS. 
We then compute the PSRR along each landless latitudinal line (being at least $2\times10^3$ km long) and average the PSRR for all days of the entire test year. The result is presented in Figure~\ref{fig:OSSE_effective_resolution}.
The GLORYS Eulerian pretraining phase leads to a very significant decrease in the model's errors across all scales. For the first day of prediction, the Lagrangian model has an effective resolution of \unit[171.4]{km}, being significantly worse than the \unit[102.9]{km} yielded by the other two models. Results are similar for the 7th day of prediction, with an effective resolution of \unit[285.8]{km} for the Lagrangian model and \unit[190.5]{km} for the other two. Lagrangian finetuning offers a small but consistent improvement at low frequencies, as shown in Figure~\ref{fig: currents plot}. We also assess the spatial scales of the directional error ratio in~\ref{apx:angle_psrr}. 

\subsection{Real-world experiments}
In the following, we adapt the different strategies defined earlier to real-world drifters. We propose to start from the Eulerian model pretrained on Glory's OSSE. In doing so, the neural network can learn meaningful synergies between its variables by being supervised on a simulation. Prior work suggest that it leads to a performance increase compared to training on simulation only \cite{archambault_james,garcia2025}. We compare the different strategies to the operational Mercator global ocean analysis and forecast version 3.6~\cite{DATAgloaf}, denominated Mercator v3.6 hereafter.

We compare the following models:
\begin{itemize}
    \item \textbf{Eulerian}. Starting from the Eulerian model trained on Glorys, we do a finetuning of this model using real-world drifters with a partial Eulerian current loss (training stages 1 and 3 in Table~\ref{tab: training stages}).
    \item \textbf{Eulerian + FineTune (FT) Lagrangian.} Starting from the Eulerian model trained on Glorys, we first do an Eulerian training on real drifters using Eulerian partial loss, and then a Lagrangian finetuning (training stages 1, 3, and 4 in Table~\ref{tab: training stages}.)
\end{itemize}

\subsubsection{Quantitative evaluation}

\begin{figure}[htbp]
    \centering
        \includegraphics[width=\linewidth]{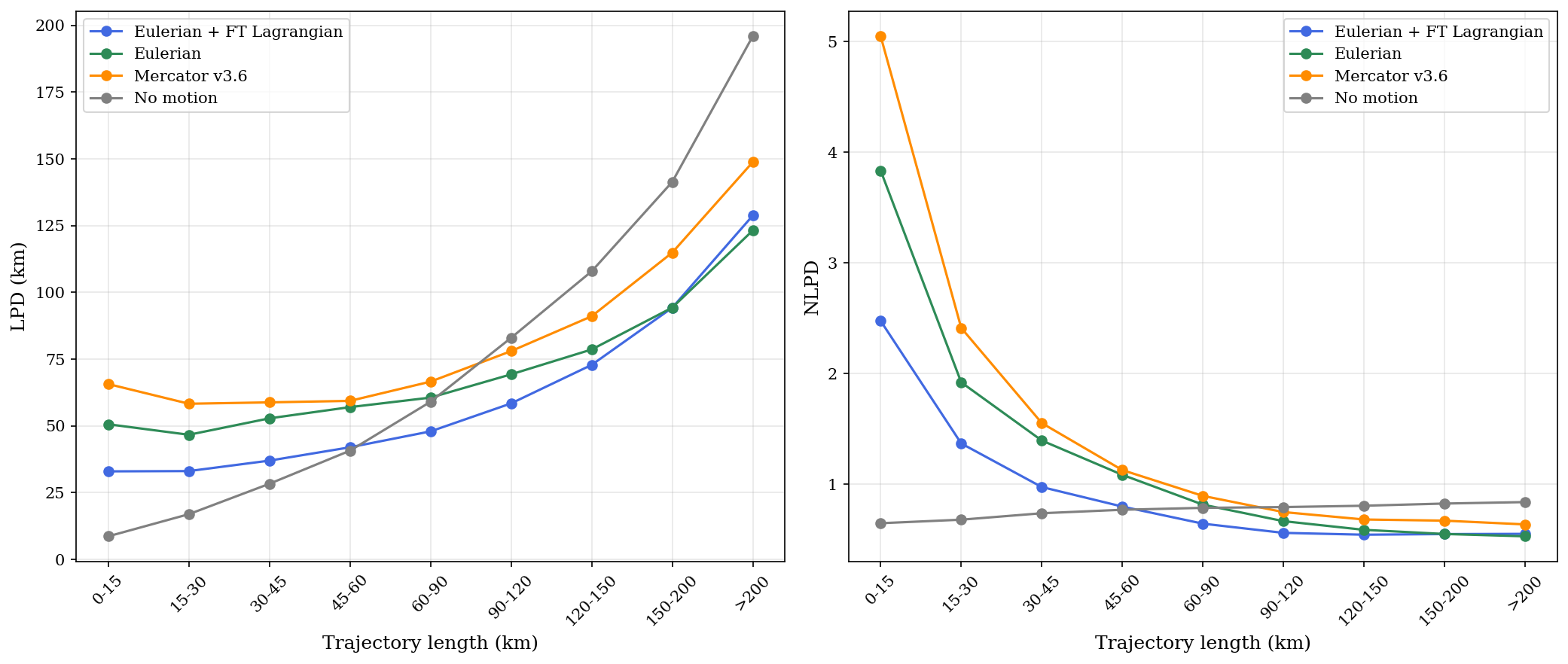}
        \caption{LPD and NLPD as a function of trajectory length on the real drifter data.}
    \label{fig: error_vs_trajectory_length_ose}
\end{figure}

We present the Lagrangian Particle Distance scores in Table~\ref{tab: drift_metrics_ose}. All scores are worse than in the idealized OSSE. This is expected, as real data is harder to predict because inputs and outputs can present more irregular observation noise, along with unresolved physical phenomena in the OSSE that take place in the real world, preventing the model from achieving the same skill regarding the target signal prediction. In particular, the full ground-truth is unavailable for the training process, which makes the fields and trajectory predictions more prone to errors. 
However, the performance increase brought by the Lagrangian fine-tuning is higher than in the OSSE experiment (14\% or LPD decrease on real data, compared to 9\% on OSSE). Additionally, it achieves an LPD 19 km lower than Mercator v3.6. In Figure \ref{fig: error_vs_trajectory_length_ose}, we see that the Lagrangian finetuned model outperforms Mercator and the Eulerian model for approximately all trajectory lengths (except for trajectory lengths above 200 km). This shows that the Lagrangian fine-tuning consistently improves the trajectory prediction.

\begin{table}[htbp]
\centering
\begin{tabular}{lcccc}
\toprule
 & \multicolumn{2}{c}{LPD} & \multicolumn{2}{c}{NLPD}   \\
\cmidrule(lr){2-3} \cmidrule(lr){4-5}  
Method & Mean & Std & Mean & Std  \\
\midrule
No motion & 80.8 & 49.1 & 0.79 & \textbf{0.20}\\
Mercator v3.6 & 80.1 & 49.5 & 0.93 & 0.68\\
Eulerian & 70.3 & 47.9 & 0.83 & 0.63\\
Eulerian + FT Lagrangian & \textbf{60.7} & \textbf{44.2} & \textbf{0.67} & 0.47\\
\bottomrule
\end{tabular}
\caption{Comparison of drift metrics for the OSE.}
\label{tab: drift_metrics_ose}
\end{table}

Following \cite{garcia2025}, we also compare the three models using Eulerian metrics. Table~\ref{tab: eul_metrics_ose} reports two metrics computed from Eulerian currents as perceived by drifters in 2024.
The first metric assesses current direction. Specifically, we compute the angle between the true and estimated current vectors and then evaluate the proportion of drifter observations for which this angular error is below 45°. This quantity, reported in the table as the Correct Ang. (Ang. for Angle), is calculated only when ground-truth current speed exceeds \unitfrac[25]{cm}{s}.

The second metric evaluates error magnitude by computing the norm of the difference between the true and predicted current vectors. This metric is calculated both for currents exceeding \unitfrac[25]{cm}{s} and for the full dataset.

Similar to Lagrangian metrics, these results indicate that both deep learning models outperform the operational Mercator Ocean v3.6 system. Lagrangian fine-tuning yields a modest improvement in directional accuracy and mean error vector norm across all drifters, although predictive skill slightly degrades at higher velocities.

\begin{table}[htbp]
\centering
\begin{tabular}{lcccccc}
\toprule
 & \multicolumn{4}{c}{Currents $> 25$~cm/s} & \multicolumn{2}{c}{All currents}   \\
\cmidrule(lr){2-5}  \cmidrule(lr){6-7}
 & 
\multicolumn{2}{c}{$\uparrow$ Correct Ang. $\%$} &
\multicolumn{2}{c}{$\downarrow$ Error Norm (cm/s)}  &
\multicolumn{2}{c}{$\downarrow$ Error Norm (cm/s)}  \\
Method & T+1 & T+7 & T+1 & T+7 & T+1 & T+7 \\
\midrule
Mercator v3.6 & 60 & 51 & 26 & 29 & 16 & 18\\
Eulerian &\textbf{81} & 65 & \textbf{19} & \textbf{24} & \textbf{12} & 15\\
Eulerian + FT Lagrangian & \textbf{81} & \textbf{67} & 20 & 25 &\textbf{12} & \textbf{14}\\
\bottomrule
\end{tabular}
\caption{Comparison of real-world drifters Eulerian metrics. Correct Ang. is the percentage of examples with an angular error below $45$°, and Error Norm is the norm of the error vector between predicted and ground truth currents.}
\label{tab: eul_metrics_ose}
\end{table}

\subsubsection{Visual comparison of trajectories}

\begin{figure}
    \centering
    \begin{subfigure}{\textwidth}
        \includegraphics[width=\linewidth]{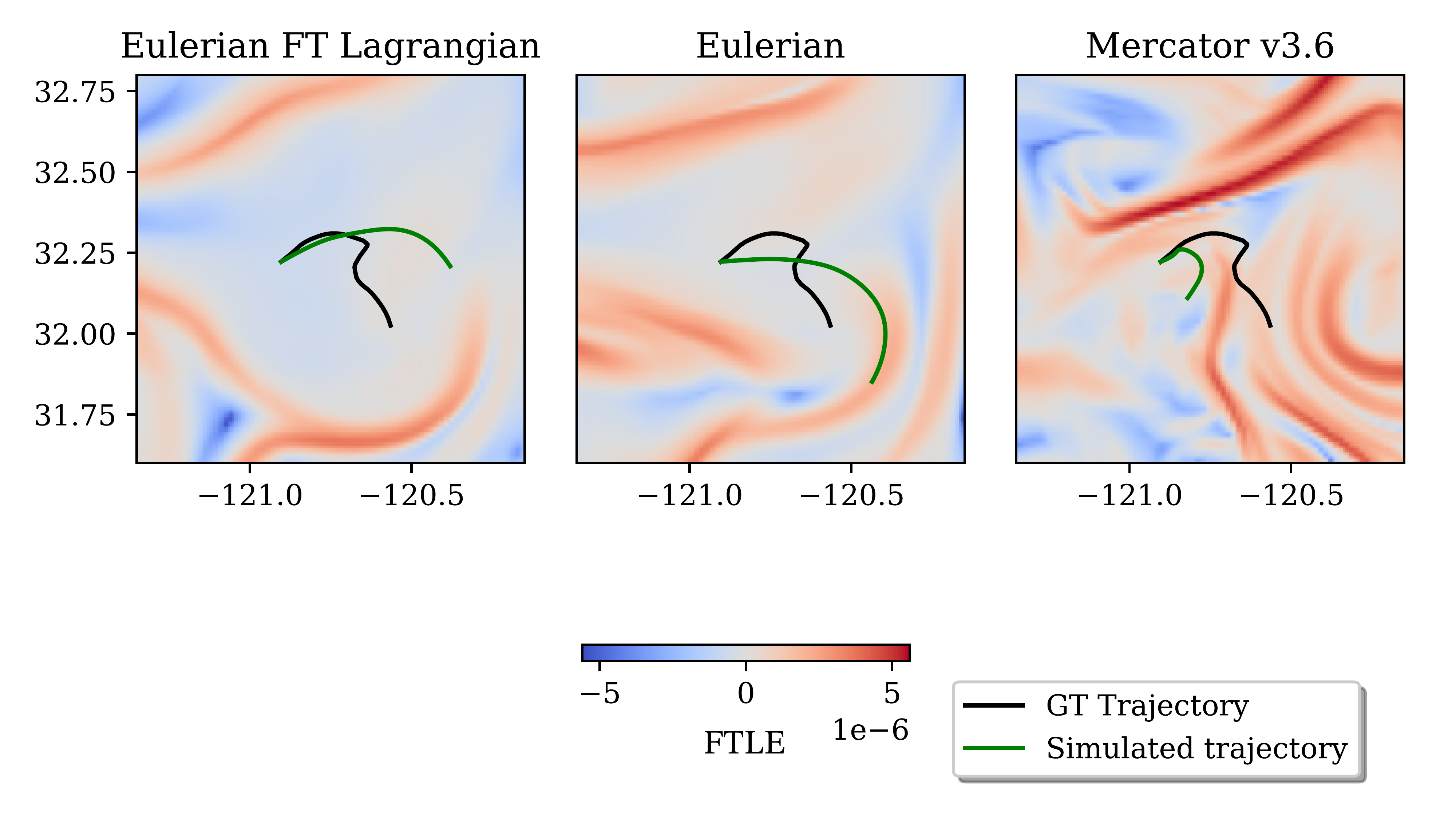}
        \subcaption{Advection of one particle and comparison to one ground-truth buoy trajectory, on the 19th of January 2024.}
    \end{subfigure}
    \begin{subfigure}{\textwidth}
        \includegraphics[width=\linewidth]{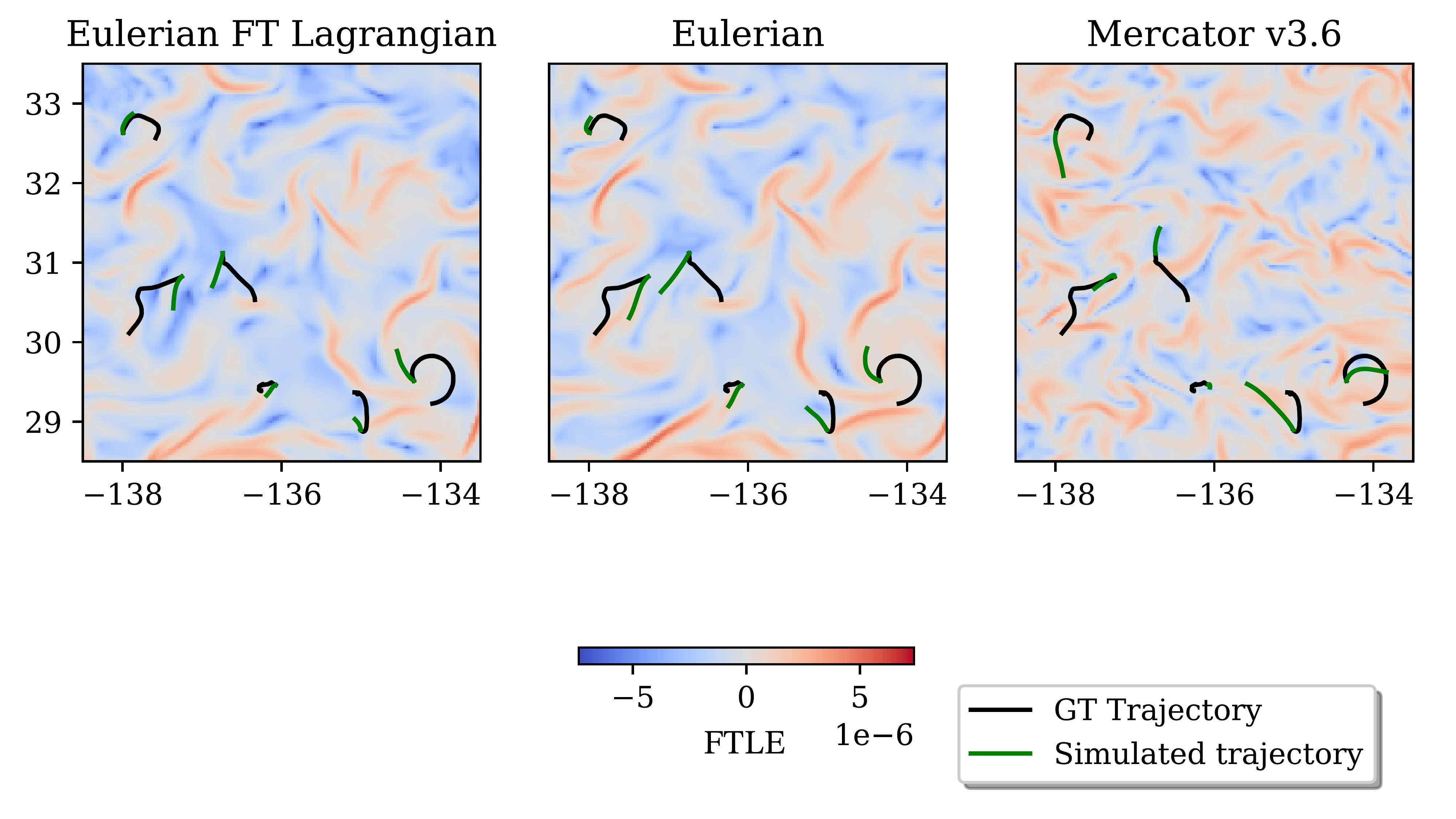}
        \subcaption{Advection of several particles and comparison to ground-truth buoy trajectories, on the 14th of June 2024.}
    \end{subfigure}
    \caption{We display Lagrangian properties of the operational DL advection maps and baseline. For solving the ODEs, we use all seven days of the operational forecasts with one-hour integration steps for the Euler solver.}
    \label{fig:OSE_qual_adv}
\end{figure}

In Figure~\ref{fig:OSE_qual_adv}, we compare the Lagrangian trajectories from each model field with those from the true drifter measurements. To do so, we plot the trajectories alongside the Finite-Time Lyapunov Exponent (FTLE), which indicates the convergence regions of the Eulerian currents. FTLE is defined as the $\log$ of the maximal eigenvalue of the right Cauchy-Green strain tensor, multiplied by $(2|T|)^{-1}$, with $T$ being the duration of the trajectory. With the Cauchy-Green strain tensor defined as $J^\top J$, with 
\begin{equation}
    J=\nabla_{r_n} \left(B^{(n)}( \hat{F}, r_n)\right)
    \label{eq: J}
\end{equation}
 where $B^{(n)}( \hat{F}, r_n)$ is the backward advection operator. We used the \textit{OpenDrift} Python module~\cite{opendrift2018} to compute the FTLE.

This qualitative comparison highlights the Eulerian FT Lagrangian model's more conservative behavior compared to the Eulerian model and the Mercator v3.6 baseline. This is visible when comparing the trajectories of simulated drifters; trajectories appear shorter for the Eulerian FT Lagrangian model. This is also visible in the FTLE fields, with the first model producing overall smoother and weaker FTLE maps.

\subsubsection{Basin-scale circulation}\label{sec: basin}

In the following section, we perform a qualitative assessment of our model's ability to reproduce basin-scale dynamics. This is an important evaluation step to ensure that the model maintains physical coherence and is capable of reproducing large-scale dynamics. To do so, we test the model's ability to replicate basin-scale convergent areas, with a focus on the North Pacific Subtropical Gyre (NPSG). The method consists in comparing remotely-sensed floating plastic accumulation measurements \cite{de_vries_automated_2026} with long-term particle simulations. 
\begin{figure}[h]
    \centering
    \begin{subfigure}{0.7\textwidth}
        \centering
        \includegraphics[width=\linewidth]{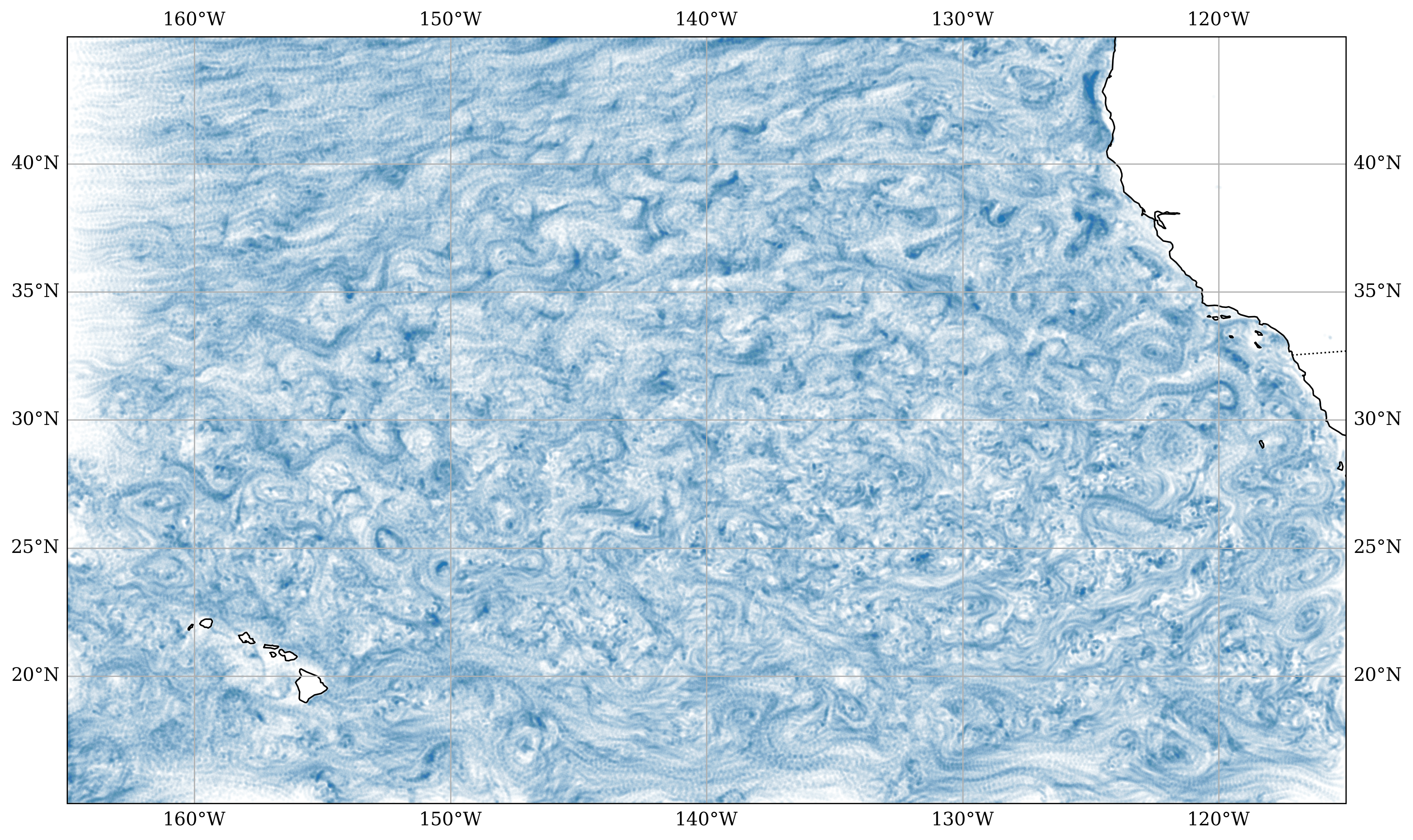}
        \subcaption{Drifter points for the first month of the drift, in January 2023.}
        \label{fig:start points}
    \end{subfigure}
    
    \begin{subfigure}{0.7\textwidth}
        \centering
        \includegraphics[width=\linewidth]{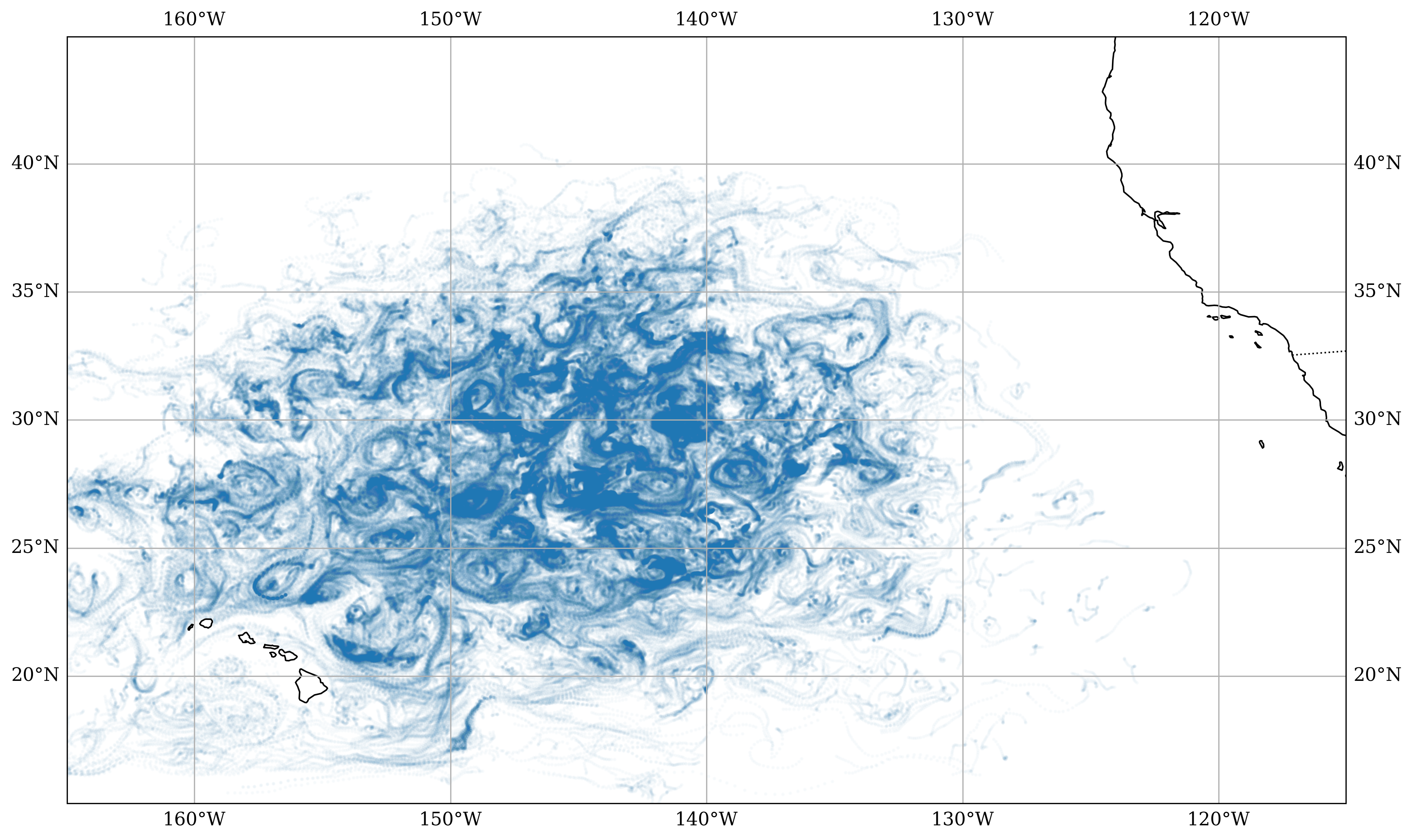}
        \subcaption{Drifter points for the last month of the drift, in December 2025.}
        \label{fig:end points}
    \end{subfigure}
    \caption{Simulated aggregation of drifters.}
\end{figure}

To this end, we release 50,000 virtual particles whose initial positions are sampled from a uniform latitude-longitude distribution over the study region. The drifters are then advected using the Eulerian + FT Lagrangian version of our model, using the currents of the first forecast day, from 1 January 2023 until 31 December 2025. We stop the advection computation when a simulated drifter exits the patch or collides with land. The difference between the particle distributions during the first and the last month of the simulation provides an intuitive representation of the formation of a convergence zone, which we use as a proxy for the North Pacific Garbage Patch (NPGP).

\begin{figure}[h]
    \centering
        \includegraphics[width=\linewidth]{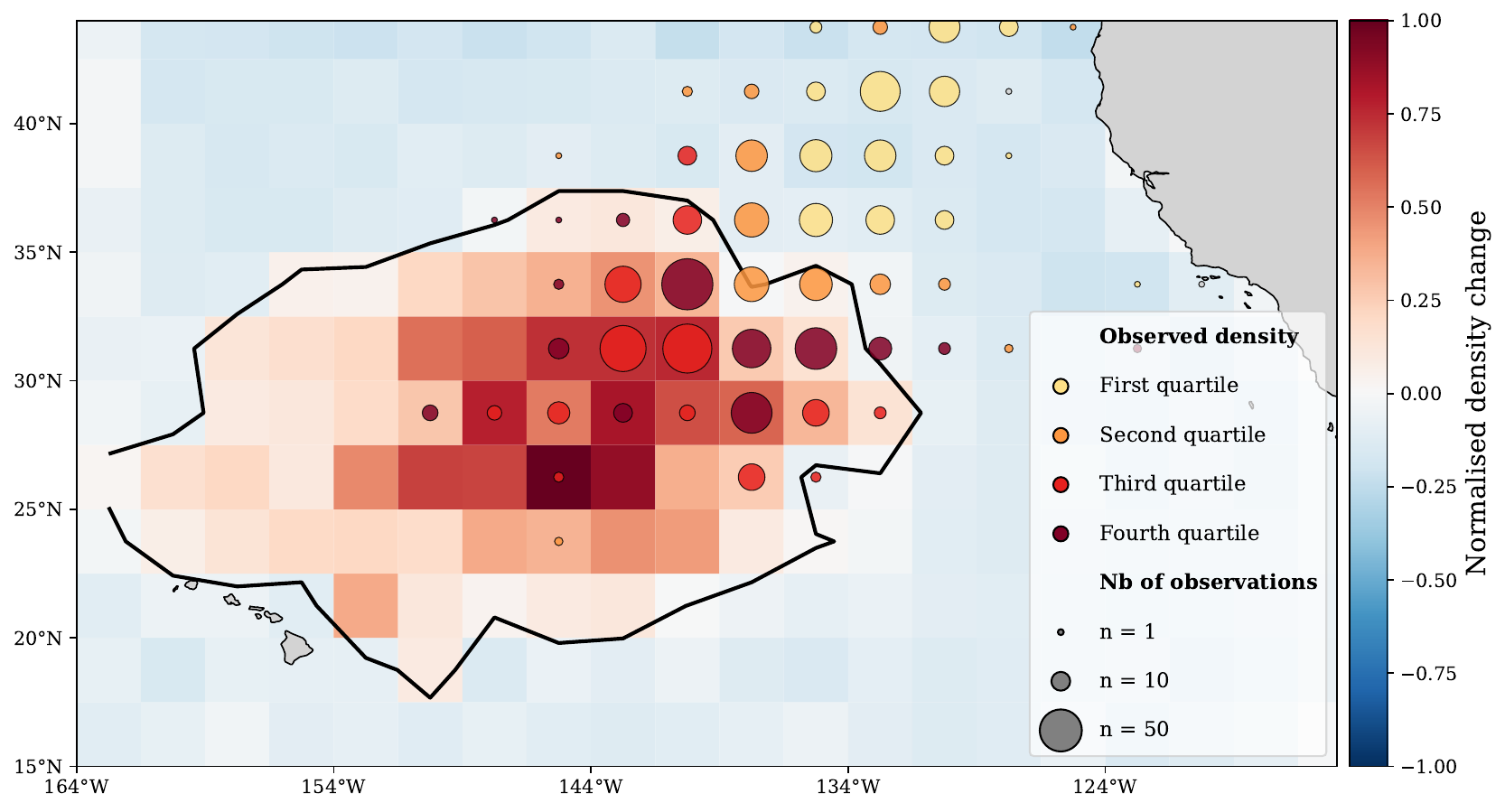}
        \caption{Comparison of modeled accumulation and dispersion zones with ADIS remotely-sensed plastic density measurements. The background field shows the normalised change in particle count between the last and first month of the drift simulation (red indicates net accumulation and blue dispersion), with the black line marking the zero-change contour. The observed plastic density is plotted as disks, where the size of the disk corresponds to the number of observations and the color represents the plastic density (with the four quartiles of the distribution).}
        \label{fig:aligned density}
\end{figure}

Figures~\ref{fig:start points} and~\ref{fig:end points} show the particle distributions during the first and last simulation months, while Figure~\ref{fig:aligned density} compares the simulated accumulation pattern with the plastic density remotely-sensed with ADIS across several field campaigns ~\cite{de_vries_automated_2026}. The simulated particle field evolves from an initially homogeneous distribution towards a well-defined accumulation region that qualitatively reproduces the location and extent of the NPGP, consistent with previous studies~\cite{lebreton2018}. The comparison with the ADIS observations shows that the simulated accumulation region overlaps well with the areas of highest observed plastic density, indicating that the model captures the dominant basin-scale transport and accumulation processes.

This experiment, nevertheless, has several limitations. First, particles are initialized from a spatially uniform distribution rather than from realistic plastic sources. Second, the simulation is restricted to the Eastern section of the NPSG, instead of representing the full North Pacific ocean basin-scale circulation and thereby preventing particles originating outside the domain from contributing to the accumulation process. Third, the three-year integration may not be sufficient for the particle distribution to reach a statistical equilibrium. Finally, the simulated accumulation pattern corresponds to December 2025, whereas the ADIS observations were collected over a period spanning from 2021 to 2024 ~\cite{de_vries_automated_2026}, limiting the temporal consistency of the comparison.

Despite these limitations, the experiment highlights the ability of the proposed model to reproduce the emergent basin-scale accumulation structure of the NPSG. It also demonstrates the computational advantage of the Lagrangian-tuned formulation, which enables the efficient advection of tens of thousands of particles using a single Eulerian forecast, rather than requiring an independent model inference for each particle trajectory.

\subsubsection{Spatial scales dynamics}

\begin{figure}
    \centering
    \begin{subfigure}{0.49\textwidth}
        \centering
        \includegraphics[width=\linewidth]{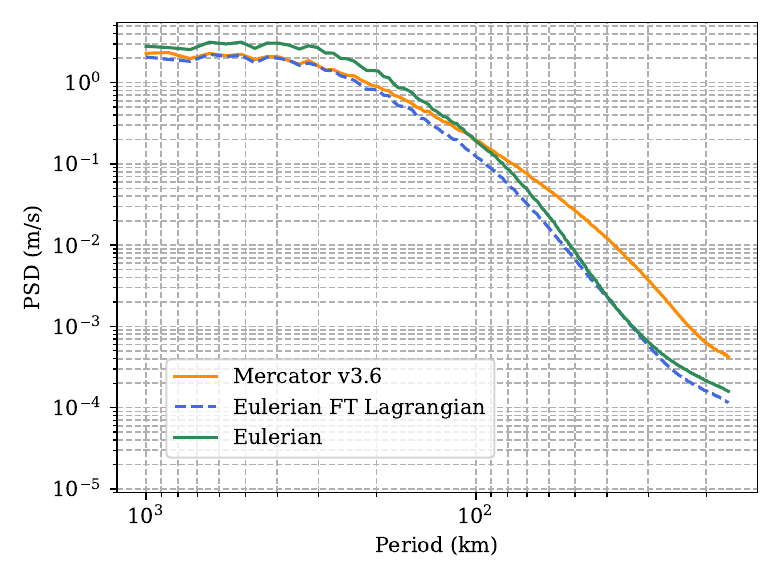}
        \caption{Current PSD for the first day of prediction.}
    \end{subfigure} 
    \begin{subfigure}{0.49\textwidth}
        \centering
    \includegraphics[width=\linewidth]{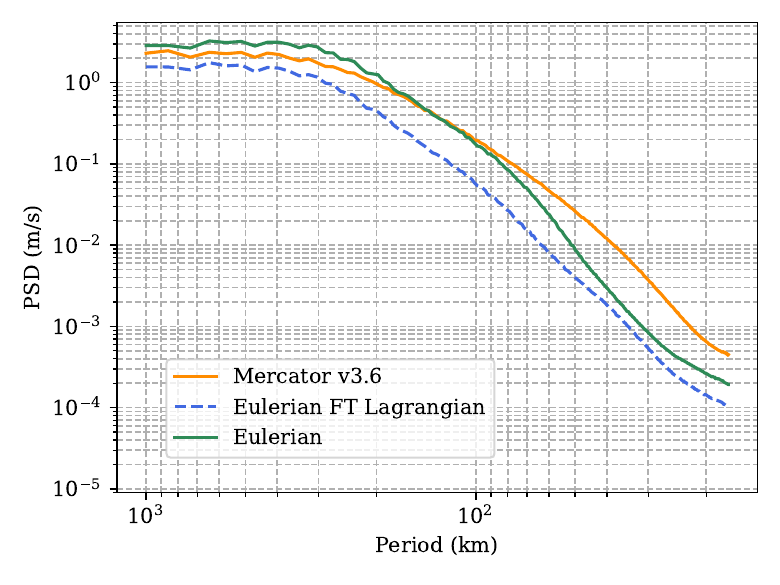}
    \caption{Current PSD for the seventh day.}
    \end{subfigure}
    \caption{For the OSE, we report the currents' PSD produced by our models and the Mercator v3.6 baseline.}
    \label{fig:OSE_PSD_comparisons}
\end{figure}

\begin{figure}
    \centering
    \begin{subfigure}{0.49\textwidth}
        \centering
        \includegraphics[width=\linewidth]{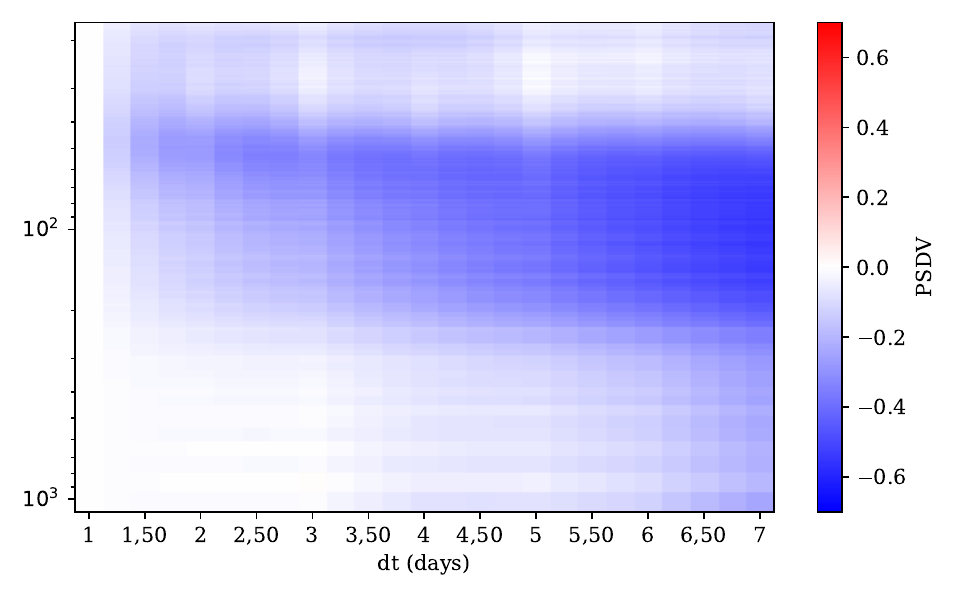}
        \caption{2024 PSDV for the OSE Eulerian FT Lagrangian model.}
    \end{subfigure}
    \begin{subfigure}{0.49\textwidth}
        \centering
    \includegraphics[width=\linewidth]{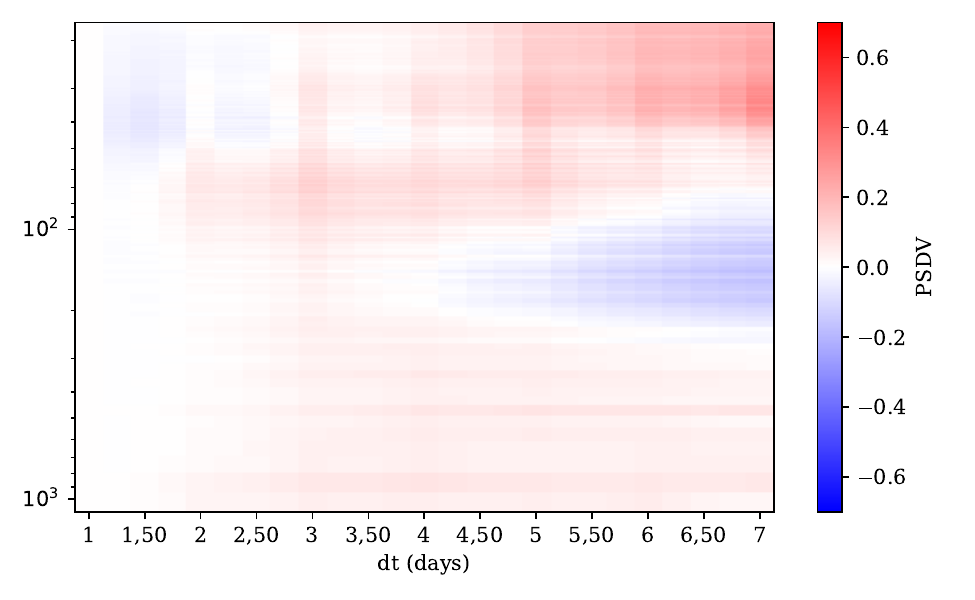}
    \caption{2024 PSDV for the OSE Eulerian model.}
    \end{subfigure}
    \caption{Relative activity of current fields on each daily timestep and with 3 sub-daily interpolations per day interval, compared to the activity of the first day. For consistency with the advection scheme, sub-daily currents are linearly interpolated from neighboring days.}
    \label{fig:OSE_PSD_norm_var_dt}
\end{figure}

Additionally, we report a comparison of fields produced by the various finetuned models and Mercator v3.6 currents. Again, we compute the PSD on the complex number representation of currents $\hat{z}=\hat{u}+i\hat{v}$. Figure~\ref{fig:OSE_PSD_comparisons} demonstrates that the \textit{Eulerian FT Lagrangian} low-scale amplitude best matches the reference Mercator v3.6 physical currents; this may be due to the less biased constraint regarding current interpolation imposed by the Advection loss, equation~\ref{eq: advection loss}. Intuitively, since the \textit{Eulerian} model has to learn daily averages of gridded drifter velocities, trajectories with high velocities will guide more strongly, as they constitute more evaluation pixels for the model, whereas the number of evaluations of the advection loss is the temporal length of the GT trajectory.

We define the Power Spectral Density Variation (PSDV)  as
\begin{equation}
    \text{PSDV}_m(\omega,t)= \text{PSD}_m(\omega,t) / \text{PSD}_m(\omega,1) - 1
\end{equation}
giving — for the model $m$ — the proportional increase of activity for timestep $t$ for wavelength $\omega$ compared to day $1$, displayed in Figure~\ref{fig:OSE_PSD_norm_var_dt}.
Current field amplitudes are the lowest with the \textit{Eulerian FT Lagrangian} model for all timesteps and spatial scales, as displayed in Figure~\ref{fig:OSE_PSD_comparisons}. From the $\text{PSDV}$, Figure~\ref{fig:OSE_PSD_norm_var_dt}, we also note a consistent decrease in all PSD amplitudes for the \textit{Eulerian FT Lagrangian} model, further demonstrating the decrease in magnitude with increasing advection uncertainty. Also in the latter figure, the temporal interpolation irregularities are more visible for the \textit{Eulerian} model, demonstrating the improvement in realism of the interpolated current for the \textit{Eulerian FT Lagrangian} model.

\section{Conclusion}

\subsection{Summary}
In this work, we present Drift Field Net (DFN), a new deep learning model for forecasting the drift of surface Lagrangian particles in the North-East Pacific. 
DFN is designed for operational use, leveraging daily satellite observations of SSH, SST, and surface winds to predict complete surface velocity fields for the following week.
The specificity of this drift predictor is that, unlike prior work, it estimates a complete flow, which can later be used to advect a large number of virtual particles, enabling efficient modeling of aggregation zones. To train this model, we introduced a new Lagrangian training strategy where a differentiable Lagrangian advection is performed using the estimated field and the initial positions of \textit{in-situ} drifters. As a result, the estimated velocity field is optimized to predict the trajectories of drifting particles, rather than pointwise Eulerian currents. The model is trained in three stages: first on an OSSE, then on observational data, and finally in a Lagrangian fine-tuning stage. 
Using \textit{in-situ} drifter observations, we demonstrate that DFN achieves a mean 7-day position error of approximately \unit[60]{km}, reducing the error by about 20 km compared with the operational Mercator global ocean analysis and forecasting system. Furthermore, the Lagrangian fine-tuning alone improves the Eulerian-trained version of DFN by an additional 10 km.

\subsection{Perspectives}

In Section~\ref{sec: basin}, we evaluated the ability of our model to reproduce the Great Pacific Garbage Patch in the North Pacific basin. However, the model was trained for operational forecasts with a prediction horizon of only seven days, which is considerably shorter than the characteristic timescales of the formation and evolution of large-scale plastic accumulation zones. 
Therefore, training DFN in \textit{hindcast} mode, by requiring it to predict long sequences of velocity fields instead of only short-term forecasts, should improve its ability to capture long-term particle trajectories and to represent the accumulation of position uncertainty over time.
Such an extension would increase computational costs and may require predicting weekly-averaged ocean currents rather than daily fields. Another promising direction would be to expand the model's spatial domain to include the entire North Pacific basin. A basin-scale framework would provide a more comprehensive representation of material transport at the ocean surface and allow for more realistic distributions of particle sources, such as coastlines, thereby improving the simulation of long-term accumulation patterns.

A second opportunity to improve the operational drift forecasts produced by the DFN framework is the incorporation of wind and wave forecasts during inference. In this work, we only used past surface wind fields, which provide limited information about the future evolution of Ekman-driven currents. Because atmospheric forcing evolves rapidly compared to ocean circulation, incorporating forecast surface winds at sub-daily temporal resolution could substantially improve the prediction of wind-driven currents. This would likely require increasing the temporal resolution of the predicted current fields accordingly. Future work should investigate strategies for jointly predicting geostrophic and Ekman currents, either by directly integrating both components within the model outputs or by applying wind-driven corrections to the predicted geostrophic currents.


\availstatement
The satellite and reanalysis datasets used in this study are publicly available. Level-3 sea surface temperature (SST) data are available from the Copernicus Marine Service (https://doi.org/10.5067/GHLDY-3S281), and Level-3 along-track sea surface height (SSH) data are available at https://doi.org/10.48670/MOI-00146. Cross-Calibrated Multi-Platform (CCMP) surface wind data are available at https://doi.org/10.56236/RSS-uv6h30. GLORYS reanalysis data are available at https://doi.org/10.48670/MOI-00016. Drifter observations are available from the Global Drifter Program at https://doi.org/10.17882/86236. The ADIS plastic debris observations are available at https://doi.org/10.4121/ddede7f5-aca5-42ae-b851-e0bbb9a2c4c2.v1. The code and data developed in this study are not publicly available.





\appendix[A] 

\appendixtitle{Angle PSRR} \label{apx:angle_psrr}

Similarly to Figure~\ref{fig:OSSE_effective_resolution}, we represent currents as complex numbers; however, for both the ground-truth and reconstructed fields, we normalize the complex numbers so that only the directional information is evaluated. We then compute the PSRR according to equation~\eqref{eq: PSRR}.
\begin{figure}[htpb]
    \centering
    \begin{subfigure}{.48\textwidth}
        \centering
        \includegraphics[width=\linewidth]{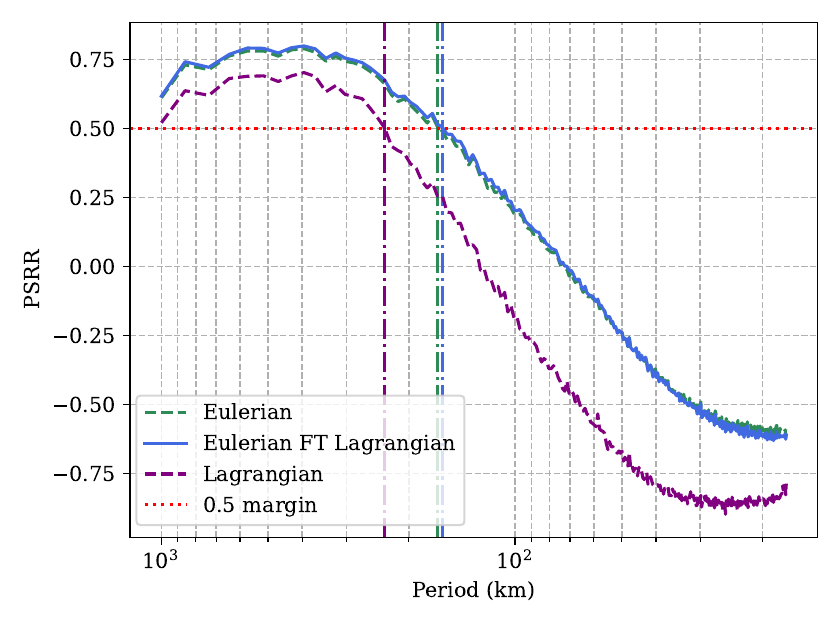}
        \caption{First day of the forecast}
    \end{subfigure}
    \begin{subfigure}{.48\textwidth}
        \centering
        \includegraphics[width=\linewidth]{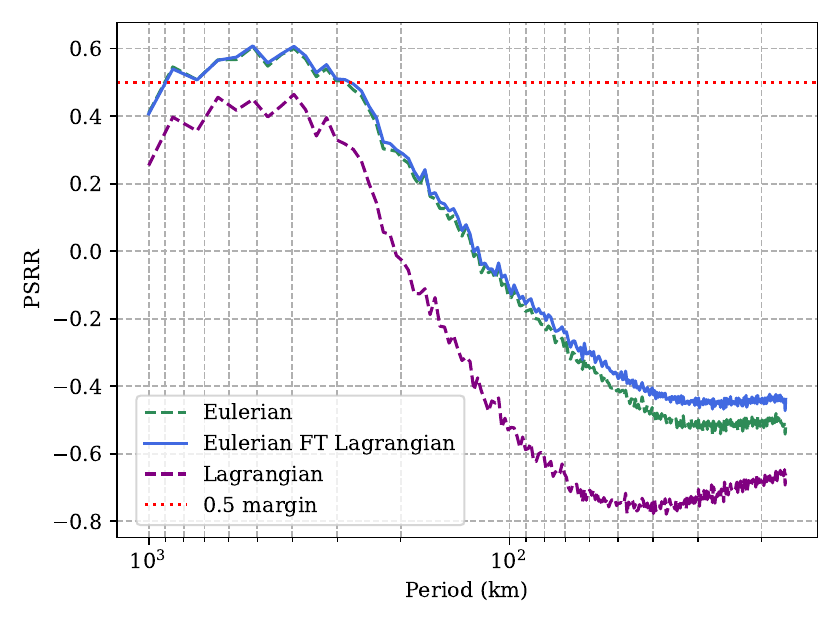}
        \caption{Seventh day of the forecast}
    \end{subfigure}
    \caption{Year 2024 average of one minus the ratio of the direction reconstruction error's PSD compared to the ground-truth (GLORYS) directions' signal. }
    \label{fig:OSSE_dir_effective_resolution}
\end{figure}

Figure~\ref{fig:OSSE_dir_effective_resolution} shows how directional errors vary after fine-tuning on the \textit{Lagrangian} loss. For the finetuned model, the figure shows directional prediction improvements at all scales for the last day of prediction, with a negligible increase for the short-wavelength error on the first day of prediction. Additionally, we note an improvement in the effective resolution on the first day of prediction for the Eulerian FT Lagrangian ($\lambda_\text{EFL}^{d=1}$) model relative to the Eulerian model ($\lambda_\text{E}^{d=1}$): $\lambda_\text{EFL}^{d=1} = 166.0 > \lambda_\text{E}^{d=1} = \unit[160.7]{km}$. Overall, Figure~\ref{fig:OSSE_dir_effective_resolution} displays a general improvement in the angular predictions for the Eulerian FT Lagrangian model, proving that Lagrangian finetuning not only diminishes predicted activity (Figure~\ref{fig:OSSE_PSD}) but also enhances directional predictions, further explaining the OSSE PSRR improvements (Figure~\ref{fig:OSSE_effective_resolution}).

\bibliographystyle{ametsocV6}
\bibliography{biblio}

@article{archambault_james,
  author = {Archambault, T. and Filoche, A. and Charantonis, A. and B\'er\'eziat, D. and Thiria, S.},
  title = {Learning Sea Surface Height Interpolation from Multi-variate Simulated Satellite Observations},
  journal = {Journal of Advances of Modelling Earth Systems},
  year = 2024,
  volume = 16,
  number = 6,
  optpages = {e2023MS004047},
  month = jun,
  optnote = {\url{https://agupubs.onlinelibrary.wiley.com/doi/10.1029/2023MS004047}}
}

@misc{wang2024,
  author = {Xiang Wang and Renzhi Wang and Ningzi Hu and Pinqiang Wang and Peng Huo and Guihua Wang and Huizan Wang and Senzhang Wang and Junxing Zhu and Jianbo Xu and Jun Yin and Senliang Bao and Ciqiang Luo and Ziqing Zu and Yi Han and Weimin Zhang and Kaijun Ren and Kefeng Deng and Junqiang Song},
  month = {2},
  title = {XiHe: A Data-Driven Model for Global Ocean Eddy-Resolving Forecasting},
    howpublished = {arXiv},
  year = {2024}
}

@article{lebreton2018,
   author = {L. Lebreton and B. Slat and F. Ferrari and B. Sainte-Rose and J. Aitken and R. Marthouse and S. Hajbane and S. Cunsolo and A. Schwarz and A. Levivier and K. Noble and P. Debeljak and H. Maral and R. Schoeneich-Argent and R. Brambini and J. Reisser},
   doi = {10.1038/s41598-018-22939-w},
   issn = {20452322},
   issue = {1},
   journal = {Scientific Reports 2018 8:1},
   month = {3},
   pages = {4666-},
   pmid = {29568057},
   publisher = {Nature Publishing Group},
   title = {Evidence that the Great Pacific Garbage Patch is rapidly accumulating plastic},
   volume = {8},
   year = {2018}
}

@article{de_vries_automated_2026,
	title = {Automated debris imaging system: calibrated global monitoring of floating marine litter using ship-based cameras and deep learning},
	volume = {8},
	copyright = {https://creativecommons.org/licenses/by/4.0/},
	issn = {2515-7620},
		doi = {10.1088/2515-7620/ae8152},
	language = {en},
	number = {7},
	urldate = {2026-07-20},
	journal = {Environmental Research Communications},
	publisher = {IOP Publishing},
	author = {de Vries, Robin and Wolter, Helen and Dalton, Sean and Klink, Doug and Romero, Mattia and Puskic, Peter and Royer, Sarah-Jeanne and Lebreton, Laurent},
	month = jul,
	year = {2026},
	pages = {075003},
}

@article{kugusheva2024,
  author = {Kugusheva, Alisa and Bull, Hannah and Moschos, Evangelos and Ioannou, Artemis and Le Vu, Briac and Stegner, Alexandre},
  title = {Ocean Satellite Data Fusion for High-Resolution Surface Current Maps},
  journal = {Remote Sensing},
  volume = {16},
  year = {2024},
  number = {7},
  article-number = {1182},
    issn = {2072-4292},
  doi = {10.3390/rs16071182}
}

@article { garcia2025,
      author = "Pierre Garcia and Inès Larroche and Amélie Pesnec and Hannah Bull and Théo Archambault and Evangelos Moschos and Alexandre Stegner and Anastase Charantonis and Dominique Béréziat",
      title = "ORCAst: Operational High-Resolution Current Forecasts",
      journal = "Artificial Intelligence for the Earth Systems",
      year = "2025",
      publisher = "American Meteorological Society",
      address = "Boston MA, USA",
      volume = "4",
      number = "4",
      doi = "10.1175/AIES-D-25-0002.1",
      pages=      "e250002",
      }

@article{fablet2023,
  author = {Ronan Fablet and Quentin Febvre and Bertrand
                  Chapron},
  doi = {10.1109/TGRS.2023.3268006},
  issn = 15580644,
  journal = {IEEE Transactions on Geoscience and Remote Sensing},
  publisher = {Institute of Electrical and Electronics Engineers
                  Inc.},
  title = {Multimodal {4DVarNets} for the Reconstruction of Sea
                  Surface Dynamics From {SST-SSH} Synergies},
  volume = 61,
  year = 2023
}

@article{martin2023,
  author = {Martin, Scott A. and Manucharyan, Georgy E. and
                  Klein, Patrice},
  title = {Synthesizing Sea Surface Temperature and Satellite
                  Altimetry Observations Using Deep Learning Improves
                  the Accuracy and Resolution of Gridded Sea Surface
                  Height Anomalies},
  journal = {Journal of Advances in Modeling Earth Systems},
  volume = 15,
  number = 5,
  optpages = {e2022MS003589},
  doi = {10.1029/2022MS003589},
    eprint = {https://agupubs.onlinelibrary.wiley.com/doi/pdf/10.1029/2022MS003589},
  optnote = {e2022MS003589 2022MS003589},
  year = 2023
}

@INPROCEEDINGS{zhangyang_simvp_2022,
  author={Gao, Zhangyang and Tan, Cheng and Wu, Lirong and Li, Stan Z.},
  booktitle={2022 IEEE/CVF Conference on Computer Vision and Pattern Recognition (CVPR)}, 
  title={SimVP: Simpler yet Better Video Prediction}, 
  year={2022},
  volume={},
  number={},
  pages={3160-3170},
  doi={10.1109/CVPR52688.2022.00317}}

@article{martin_deep_2024,
	title = {Deep {Learning} {Improves} {Global} {Satellite} {Observations} of {Ocean} {Eddy} {Dynamics}},
	volume = {51},
	copyright = {© 2024. The Author(s).},
	issn = {1944-8007},
		doi = {10.1029/2024GL110059},
	language = {en},
	number = {17},
	urldate = {2024-10-02},
	journal = {Geophysical Research Letters},
	author = {Martin, Scott A. and Manucharyan, Georgy E. and Klein, Patrice},
	year = {2024},
	note = {\_eprint: https://onlinelibrary.wiley.com/doi/pdf/10.1029/2024GL110059},
	pages = {e2024GL110059},
}

@article{martin2025,
author = {Martin, Scott A. and Manucharyan, Georgy E. and Klein, Patrice},
title = {Generative Data Assimilation for Surface Ocean State Estimation From Multi-Modal Satellite Observations},
journal = {Journal of Advances in Modeling Earth Systems},
volume = {17},
number = {8},
pages = {e2025MS005063},
doi = {https://doi.org/10.1029/2025MS005063},
eprint = {https://agupubs.onlinelibrary.wiley.com/doi/pdf/10.1029/2025MS005063},
note = {e2025MS005063 2025MS005063},
year = {2025}
}

@article{resac,
  author = {Thiria, S. and Sorror, C. and Archambault, T. and
                  Charantonis, A. and Béréziat, D. and Mejia, C. and
                  Molines, J.-M.  and Crepon, M.},
  title = {Downscaling of ocean fields by fusion of
                  heterogeneous observations using Deep Learning
                  algorithms},
  year = 2023,
  volume = 182,
  journal = {Ocean Modeling}
}

@misc{madec2017nemo,
  title = {{NEMO} ocean engine},
  author = {Madec, G. and Bourdall{\'e}-Badie, R. and Bouttier,
                  P.-A.  and Bricaud, C. and Bruciaferri, D. and
                  Calvert, D. and Chanut, JSSH. and Clementi, E. and
                  Coward, A. and Delrosso, D. and others},
  year = 2017,
  howpublished = {Scientific Notes of Climate Modelling Center}
}

@article { botvynko2025,
      author = "Daria Botvynko and Carlos Granero-Belinchon and Simon van Gennip and Abdesslam Benzinou and Ronan Fablet",
      title = "Neural Prediction of Lagrangian Drift Trajectories on the Sea Surface",
      journal = "Artificial Intelligence for the Earth Systems",
      year = "2025",
      publisher = "American Meteorological Society",
      address = "Boston MA, USA",
      volume = "4",
      number = "3",
      doi = "10.1175/AIES-D-24-0052.1",
      pages=      "e240052",
      }

@article{opendrift2018,
    author = {Knut-Frode Dagestad AND Johannes Röhrs AND Øyvind Breivik AND Bjørn Ådlandsvik},
    title = {OpenDrift v1.0: a generic framework for trajectory modelling},
    journal = {Geoscientific Model Development, 11}, 
    pages = {1405-1420},
    doi= {https://doi.org/10.5194/gmd-11-1405-2018},
    year = "2018"
}

@article{delandmeter_parcels_2019,
	title = {The {Parcels} v2.0 {Lagrangian} framework: new field interpolation schemes},
	volume = {12},
	issn = {1991-959X},
	shorttitle = {The {Parcels} v2.0 {Lagrangian} framework},
		doi = {10.5194/gmd-12-3571-2019},
	language = {English},
	number = {8},
	urldate = {2026-05-20},
	journal = {Geoscientific Model Development},
	publisher = {Copernicus GmbH},
	author = {Delandmeter, Philippe and van Sebille, Erik},
	month = aug,
	year = {2019},
	pages = {3571--3584},
}

@article{garcia_glofm_2026,
	title = {{GloFM}: {A} {GLORYS} {Flow}-{Matching} {Emulator} for {Spatio}-{Temporal} {Ocean} {Data} {Assimilation}},
	isbn = {978-989-758-804-4},
	shorttitle = {{GloFM}},
	urldate = {2026-05-20},
	author = {Garcia, Pierre and Archambault, Théo and Béréziat, Dominique and Charantonis, Anastase},
	month = may,
	year = {2026},
	pages = {411--418},
    journal = {VISAPP}
}

@article{champenois_predicting_2026,
	chapter = {Artificial Intelligence for the Earth Systems},
	title = {Predicting {Ocean} {Flow} from {Observed} {Lagrangian} {Trajectories} via {Bayesian} {Optimization} in the {Latent} {Space} of an {Autoencoder}-{Based} {Reduced}-{Order} {Model}},
	volume = {-1},
	issn = {2769-7525},
		doi = {10.1175/AIES-D-25-0124.1},
	language = {EN},
	number = {aop},
	urldate = {2026-05-19},
	journal = {Artificial Intelligence for the Earth Systems},
	publisher = {American Meteorological Society},
	author = {Champenois, Bianca and Sapsis, Themistoklis P.},
	month = apr,
	year = {2026},
}

@article{tian_prediction_2025,
	title = {Prediction of surface drifter trajectories in the {South} {China} sea using deep learning},
	volume = {15},
	copyright = {2025 The Author(s)},
	issn = {2045-2322},
		doi = {10.1038/s41598-025-20143-1},
	language = {en},
	number = {1},
	urldate = {2026-05-20},
	journal = {Scientific Reports},
	publisher = {Nature Publishing Group},
	author = {Tian, Chuan and Wang, Ying and Xia, Ruixue and Liang, Yun and Song, Yuanjie and Xu, Dazhen and Xu, Xiaoyang and Wang, Chen},
	month = oct,
	year = {2025},
	pages = {36361},
}

@article{della_cioppa_predicting_2025,
	title = {Predicting oceanic {Lagrangian} trajectories with hybrid space-time {CNN} architecture},
		doi = {10.5194/egusphere-2025-1136},
	language = {English},
	urldate = {2026-05-20},
	journal = {EGUsphere},
	publisher = {Copernicus GmbH},
	author = {Della Cioppa, Lorenzo and Buongiorno Nardelli, Bruno},
	month = may,
	year = {2025},
	pages = {1--18},
}

@misc{DATAmur,
  title = {{GHRSST} Level 4 {MUR} 0.25 deg Global Foundation Sea
                  Surface Temperature Analysis (v4.2) [Dataset]},
  year = 2019,
  publisher = {NASA},
  author = {NASA/JPL},
  note = {\url{https://doi.org/10.5067/GHGMR-4FJ04}}
}

@misc{DATAsstl3,
	title = {{GHRSST} {NOAA}/{STAR} {ACSPO} v2.81 0.02 degree {L3S} {Daily} {Dataset} from {LEO} {Satellites}},
    year = 2023,
    type={Dataset},
	urldate = {2025},
	journal = {Physical Oceanography Distributed Active Archive Center (PO.DAAC)},
    doi = {10.5067/GHLDY-3S281},
    note = {\url{https://doi.org/10.5067/GHLDY-3S281}},
    editor={PO.DAAC},
    author={PO.DAAC}    
}

@misc{DATAdrifters,
  title = {Copernicus Marine In Situ - Global Ocean-Delayed Mode in situ Observations of surface (drifters, {HFR}) and sub-surface (vessel-mounted {ADCP}s) water velocity},
  type = {Dataset},
  note = {\url{https://doi.org/10.17882/86236}},
  doi = {10.17882/86236},
  year = {2024},
  editor = {SEANOE},
  author = {SEANOE}
}

@misc{DATAourOSSE,
  author = {Théo Archambault},
  title = {20 years {OSSE} and {OSE} of {SSH} and {SST} data in the {G}ulf {S}tream area [Dataset]},
  month = jan,
  year = 2024,
  publisher = {Zenodo},
  doi = {10.5281/zenodo.10551897},
  note = {\url{https://doi.org/10.5281/zenodo.10551897}}
}

@misc{DATAL3tracks,
  title = {Global Ocean Along-Track {L3} Sea Surface Heights
                  Reprocessed (1993-Ongoing) Tailored For Data
                  Assimilation [Dataset]},
  year = 2021,
  publisher = {Mercator Ocean International},
  author = {CMEMS},
  note = {\url{https://doi.org/10.48670/MOI-00146}}
}

@misc{DATAglorys,
  title = {Global Ocean Physics Reanalysis [Dataset]},
  year = 2020,
  publisher = {Mercator Ocean International},
  author = {CMEMS},
  note = {\url{https://doi.org/10.48670/moi-00021}}
}

@misc{DATAgloaf,
  title = {Global Ocean Physics Analysis and Forecast [Dataset]},
  year = 2025,
  publisher = {Mercator Ocean International},
  author = {CMEMS},
  note = {\url{https://doi.org/10.48670/moi-00016}}
}

@misc{DATAccmp,
title = "RSS Cross-Calibrated Multi-Platform (CCMP) 6-hourly ocean vector wind analysis on 0.25 deg grid, Version 2.",
type = "Dataset",
doi = "10.56236/RSS-uv6h30",
author = "Mears, C. and Lee, T. and Ricciardulli, L. and Wang, X. and Wentz, F.",
year = "2022",
editor="Remote Sensing Systems",
 
}

@misc{ODC2021a,
title = "Ocean data challenges: 2021a SSH mapping OSE",
type = "repository",
doi = "https://doi.org/10.5281/zenodo.5511905",
author = "Maxime Ballarotta AND Samy Metref AND A Albery AND Emmanuel Cosme AND Maxime Beauchamp AND Florian Le Guillou (2021).",
year = "2021"
}

\end{document}